\documentclass{fairmeta}
\usepackage{amsmath,amsfonts,bm}

\usepackage{amsthm}
\theoremstyle{remark}
\newtheorem*{remark}{Remark}
\theoremstyle{plain}
\newtheorem{proposition}{Proposition}[section]

\newtheorem{theorem}{Theorem}
\newtheorem{lemma}[theorem]{Lemma}

\usepackage{xspace}
\usepackage{url}
\usepackage{subcaption}
\usepackage[inline]{enumitem}
\usepackage[most]{tcolorbox}
\usepackage{booktabs,tabularx}
\usepackage{xcolor,colortbl}
\usepackage{float}
\floatstyle{ruled}
\newfloat{algorithm}{tbp}{loa}
\floatname{algorithm}{Algorithm}
\floatstyle{plain}
\usepackage{multirow}
\usepackage{caption}
\usepackage{graphicx}

\usepackage{adjustbox}
\usepackage{wrapfig}
\usepackage{amssymb}
\usepackage{hyperref}
\usepackage[capitalize,nameinlink]{cleveref}
\crefname{algorithm}{algorithm}{algorithms}
\Crefname{algorithm}{Algorithm}{Algorithms}

\definecolor{promptbg}{HTML}{F6F8FA}
\definecolor{promptborder}{HTML}{D0D7DE}
\definecolor{prompttitlebg}{HTML}{EEF2F6}
\definecolor{opdgreen}{RGB}{28,120,70}
\definecolor{opdred}{RGB}{180,55,55}
\definecolor{oursblue}{RGB}{232,242,252}
\definecolor{scalegray}{RGB}{244,245,247}

\newcommand{\methodname}{\textsc{IT-RFT}}
\title{Inference-Time Weak-to-Strong Transfer for Efficient Distillation}

\author[1, \star]{Yuxiao Yang}
\author[1]{Shangzhe Li}
\author[2]{Tianrun Yu}
\author[2]{Kaixiang Zhao}
\author[2,\dagger]{Taylor W. Killian}
\author[1,\ddagger]{Weitong Zhang}
\contribution[\star]{Project lead, \email{yxyang@unc.edu}}
\contribution[\dagger]{\email{tkillian@cs.byu.edu}}
\contribution[\ddagger]{\email{weitongz@unc.edu}}

\affiliation[1]{University of North Carolina at Chapel Hill}
\affiliation[2]{Brigham Young University}
\begin{document}
\abstract{
Weak-to-strong on-policy distillation can improve larger language models using guidance from smaller teachers, but identifying useful guidance often requires repeated student training.
We study inference-time weak-to-strong transfer to evaluate candidate guidance through student responses before updating model parameters.
Using the closed-form policy of a KL-regularized distillation objective, we apply weak-teacher guidance directly during generation with frozen models.
Its performance follows the trends of trained Direct-OPD students across teacher checkpoints, enabling efficient comparison of supervision sources.
We then develop inference-time-guided rejection fine-tuning (\methodname{}), which uses verified responses from this policy to train a student that subsequently runs on its own.
Across math, code, and instruction following, \methodname{} achieves competitive or higher performance than OPD baselines while requiring up to $7.3\times$ fewer GPU-hours in mixed-domain distillation.
The same weak teachers transfer across student scales, and their student-conditioned responses yield stronger fine-tuning results than responses generated directly by more capable models.
}

\maketitle

\section{Introduction}
\label{sec:intro}

Weak-to-strong on-policy distillation (OPD) provides a way to improve large language models using guidance from smaller teachers \citep{DirectOPD,W2SOPD}.
By evaluating teacher feedback on student-generated prefixes, OPD transfers useful behavior on the student's own generation distribution \citep{gkd,opd2}.
In the weak-to-strong setting, this feedback can come from the policy shift between a small base model and its reinforcement learning (RL)-trained counterpart.
The shift captures improvements learned through RL and can guide a larger student even when the teacher has lower overall capability.

The effectiveness of this transfer depends on \emph{which weak-teacher guidance the student should follow}.
A teacher's standalone performance does not establish how useful its guidance will be to a stronger student.
Different checkpoints can encode different improvements, and different domains may call for different teachers.
Existing methods support sequential transfer, domain routing, or the combination of multiple teacher signals \citep{DirectOPD,W2SOPD,delta-mopd}.
Assessing these choices through downstream OPD requires student rollouts, teacher scoring, and parameter updates for each candidate configuration.
Teacher selection can therefore involve several distillation runs before a useful student is obtained.

Inference-time methods offer a way to explore model behavior without repeatedly updating its parameters by adaptive allocation of inference computation to improve reason quality~\citep{selfconsistency,snell2025scaling, icpo}.
Such an decoding-time guidance can also transfer preferences from smaller models into a larger model's generation distribution \citep{DExperts,Proxytuning}.
These developments suggest a change to the weak-to-strong training procedure: evaluate teacher guidance through the responses it elicits at inference, then use the resulting responses to train the student.
Learning from generated reasoning already supports effective post-training \citep{rft,star}; here, the opportunity is to make inference-time evaluation of weak-teacher guidance part of the same procedure.

We study \emph{inference-time weak-to-strong transfer} as a means of selecting and applying guidance before student training.
Our starting point is the closed-form policy of the state-wise KL-regularized distillation objective.
This policy combines the student's reference distribution with the relative preferences of two weak teachers and can be sampled directly with frozen models.
The construction connects decoding-time logit guidance \citep{DExperts,mitchell2024eft,Proxytuning} with the proxy teacher used in W2S-OPD \citep{W2SOPD}.
Across teacher checkpoints in math, code, and instruction following, its validation performance follows the trends of trained Direct-OPD students at a fraction of the computational cost.
This correspondence enables candidate guidance to be assessed before a student-training run.
A support-aware gate further improves the reliability of generation by suppressing positive guidance increments for tokens unsupported by both teachers.

We also characterize the realizability and optimization of this policy in a linear-softmax model (\Cref{sec:theory}).
When the student's features cover the weak teacher's centered logit change, the closed-form policy is exactly realizable by a student output head, and ordinary head gradient descent converges geometrically to that policy.
This coverage condition explains how a weak teacher's learned change can be evaluated directly within a richer student model before training.

\begin{figure}[t]
    \centering
    \includegraphics[width=\linewidth]{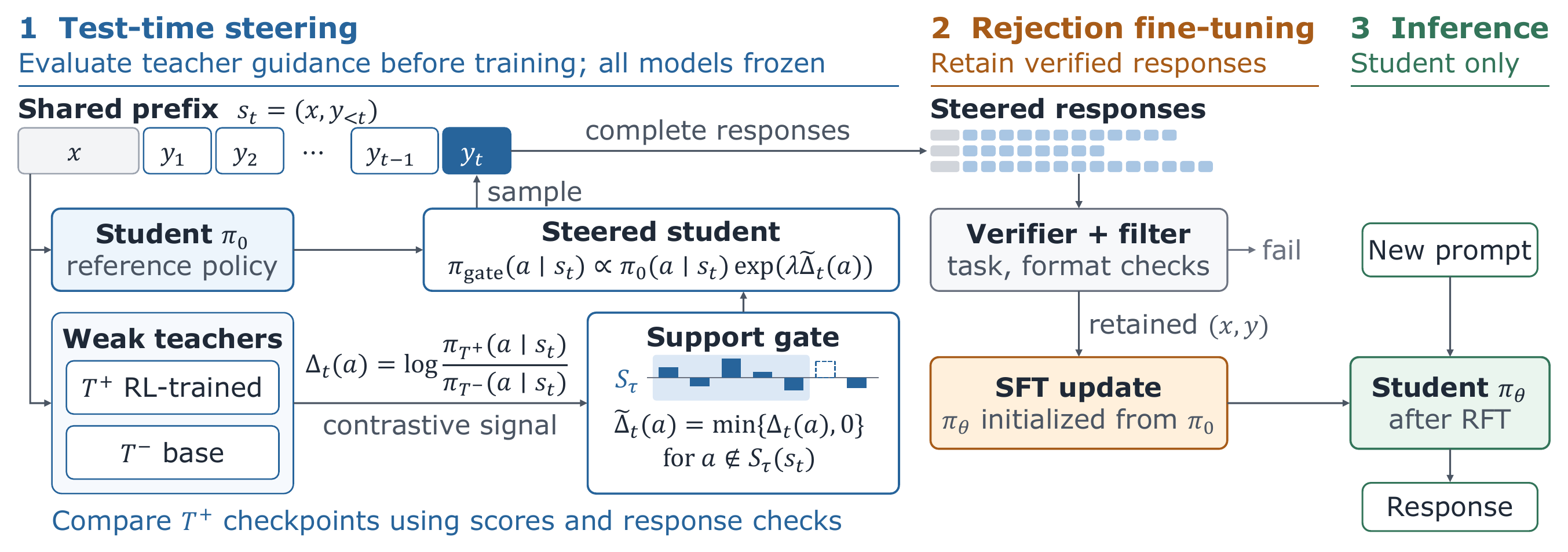}
    \caption{\textbf{Overview of \methodname{}.} Frozen-model inference evaluates guidance and generates verified responses for RFT. Blue denotes frozen models, orange student updates, and green student-only inference.}
    \label{fig:pipeline}
\end{figure}

We introduce inference-time-guided rejection fine-tuning (\methodname{}) to transfer these responses into the student's parameters.
The procedure generates student-conditioned trajectories with the selected guidance, retains those that pass verification and output checks, and applies supervised fine-tuning.
For multiple domains, accepted responses generated with the corresponding domain teachers are pooled to train one student.
Inference-time computation thus serves both to evaluate supervision and to construct the data used for transfer.
After training, the student generates independently of the weak teachers.
\Cref{fig:pipeline} summarizes the procedure.

Our experiments evaluate \methodname{} across math, code, and instruction following.
It achieves competitive or higher performance than weak-to-strong OPD baselines, with $6.8$--$7.3\times$ fewer GPU-hours in mixed-domain training.
The same 1.7B teacher pair improves both 4B and 14B students.
On math, student-conditioned responses also yield better RFT results than responses generated directly by the evaluated stronger teachers, despite providing fewer verified trajectories.
These results connect the usefulness of weak-teacher guidance to the responses it induces in the student.

Our contributions are:
\begin{itemize}[leftmargin=*]
    \item \textbf{Inference-time evaluation of weak-teacher guidance.}
    We show that directly sampling the state-wise distillation policy captures the performance trends of trained Direct-OPD students across teacher checkpoints.
    This provides a practical basis for comparing guidance before committing to student training.

    \item \textbf{Efficient weak-to-strong transfer through \methodname{}.}
    We combine support-aware inference-time guidance with rejection fine-tuning to transfer verified responses into a single student.
    The procedure supports both single-domain and mixed-domain learning and achieves competitive or higher accuracy than the OPD baselines at substantially lower computational cost.

    \item \textbf{Student-conditioned supervision across model scales.}
    We show that responses generated by guiding the student can be more effective for RFT than responses from stronger standalone teachers.
    The same weak-teacher pair also transfers to different student sizes, allowing its learned guidance to be reused across models.
\end{itemize}

\section{Related Work}
\label{sec:related}

\paragraph{On-Policy and Weak-to-Strong Distillation.}
On-policy distillation (OPD) obtains teacher supervision at student-generated prefixes, addressing the mismatch between fixed training sequences and the student's generation distribution \citep{gkd}. Weak-to-strong distillation applies this recipe when the teacher is less capable overall, using improvements learned by weaker models to guide a stronger student \citep{DirectOPD,W2SOPD}. Direct-OPD interprets the log-probability ratio between a teacher before and after RL as a dense reward for KL-regularized student optimization \citep{DirectOPD}. W2S-OPD adds weak-model logit differences to a reference student's logits to construct a proxy teacher, including combinations of multiple capability directions \citep{W2SOPD}. Waypoint-OPD uses an intermediate teacher checkpoint to measure agreement between early and late policy shifts and gate the transferred signal \citep{waypointOPD}. These formulations expose teacher and checkpoint choices that affect transfer. A teacher's standalone score does not directly measure the value of its guidance to a particular student, motivating evaluation through student-conditioned generation. Our work assesses candidate checkpoints with a fixed student, studies how inference-time performance tracks Direct-OPD, and consolidates the selected guidance through verified responses.

\paragraph{Inference-Time Guidance with Small Models.}
DExperts combines a pretrained model with expert and anti-expert predictions for controllable generation \citep{DExperts}. Contrastive Decoding contrasts large- and small-model likelihoods under a plausibility constraint \citep{ContrastiveDecoding}. Emulated fine-tuning and proxy-tuning transfer behavioral changes learned by smaller models to larger models through logit arithmetic, while retaining the larger model's parameters \citep{mitchell2024eft,Proxytuning}. Weak-to-Strong Search uses small tuned and untuned models to score candidate continuations from a frozen larger model \citep{W2SSearch}. ThinkLogit applies logit differences from small reasoning models to elicit long reasoning in a larger model \citep{zhang2025thinklogit}. More recently, CritICL transfers weaker-model failure patterns through critique-based in-context examples \citep{criticL}. These methods establish several ways to make weak-model information useful during inference. We use the closed-form policy associated with the state-wise W2S objective to compare teacher guidance, and introduce support-aware gating to control the promotion of tokens with negligible teacher probability.

\paragraph{Test-Time Scaling and Adaptation.}
Test-time scaling studies how additional sampling, search, and iterative refinement can improve model responses. Self-consistency samples multiple reasoning paths and aggregates their answers \citep{selfconsistency}. \citet{snell2025scaling} study verifier-guided search and adaptive response refinement, showing that the effectiveness of additional computation depends on problem difficulty. In-Context Policy Optimization (ICPO) refines responses using previous attempts and their rewards in context while keeping model parameters fixed \citep{icpo}. Test-Time Reinforcement Learning (TTRL) uses majority voting to estimate rewards on unlabeled test data and updates model parameters through RL \citep{ttrl}. These approaches illustrate distinct uses of inference computation: selecting answers, refining a response distribution, and adapting parameters. We use inference-time computation to assess weak-teacher transfer and generate training responses, keeping the student fixed throughout this stage. This makes candidate guidance directly evaluable before the subsequent offline fine-tuning run.

\paragraph{Learning from Generated and Verified Responses.}
Generated responses can also supply supervision for model training. STaR iterates between rationale generation and fine-tuning on rationales that yield correct answers, using answer-conditioned rationalization to recover additional examples \citep{star}. Rejection fine-tuning collects correct sampled reasoning paths for supervised learning \citep{rft}, and ReST separates sample generation from offline policy improvement so that generated data can support repeated updates \citep{rest}. The s1 approach combines supervised learning on a curated set of reasoning traces with inference-time budget forcing \citep{s1}. CoDIT constructs instruction-tuning data through contrastive decoding between post-trained and pretrained models \citep{CODIT}. Building on this connection between inference and training, \methodname{} generates responses from the student conditioned on selected weak-teacher guidance and fine-tunes on the verified trajectories. The study links teacher evaluation to the data-generation policy and tests whether these student-conditioned responses improve transfer across domains and student sizes. After RFT, the resulting student generates without auxiliary teacher calls.

\section{Preliminaries}
\label{sec:prelim}

\paragraph{On-Policy Distillation.}
On-policy distillation (OPD) transfers knowledge from a teacher policy $\pi_T$ to a student policy $\pi_\theta$ using student-generated trajectories.
Given a prompt $x \sim \mathcal{D}$ and a response $y \sim \pi_\theta(\cdot \mid x)$, let $s_t=(x,y_{<t})$ denote the prefix at decoding step $t$.
OPD typically minimizes the reverse KL divergence between the student and teacher distributions:
\begin{align}
    \mathcal{L}_{\mathrm{OPD}}(\theta)
    &=
    \mathbb{E}_{x\sim\mathcal{D},\,y\sim\pi_\theta(\cdot\mid x)}
    \left[
    {\textstyle\sum_{t=1}^{|y|}}
    D_{\mathrm{KL}}\left(
    \pi_\theta(\cdot\mid s_t)
    \,\|\,\pi_T(\cdot\mid s_t)
    \right)
    \right]. \notag
\end{align}
Treating a student-visited prefix $s_t$ as fixed, the negative reverse KL gradient is
\begin{align}
    -\nabla_\theta D_{\mathrm{KL}}
    \left(
    \pi_\theta(\cdot\mid s_t)
    \,\|\,\pi_T(\cdot\mid s_t)
    \right)
    &=
    \mathbb{E}_{a\sim\pi_\theta(\cdot\mid s_t)}
    \left[
    A_t(a)\nabla_\theta
    \log\pi_\theta(a\mid s_t)
    \right], \notag
\end{align}
where
$A_t(a)=\log\pi_T(a\mid s_t)-\log\pi_\theta(a\mid s_t)$
provides dense token-level supervision.

\paragraph{Weak-to-Strong Distillation.}
Weak-to-strong (W2S) distillation methods \citep{DirectOPD,W2SOPD} instead exploit the behavioral difference between two weaker policies, a positive teacher $\pi_{T^+}$ and a negative teacher $\pi_{T^-}$.
The contrastive signal is defined as
\begin{align}
    \Delta_t(a)
    &=
    \log\pi_{T^+}(a\mid s_t)
    -
    \log\pi_{T^-}(a\mid s_t).
    \label{eq:weak_signal}
\end{align}
This signal measures the relative change in token preferences between the two policies.
In our experiments, $T^-$ is a small base model and $T^+$ is a checkpoint obtained by training that model with RL.

Given a fixed student reference policy $\pi_0$, we consider the following state-wise KL-regularized objective:
\begin{align}
    \mathcal{J}_{\mathrm{W2S}}(\theta;s_t)
    &=
    \mathbb{E}_{a\sim\pi_\theta(\cdot\mid s_t)}
    \left[\Delta_t(a)\right]
    \quad -
    \beta D_{\mathrm{KL}}
    \left(
    \pi_\theta(\cdot\mid s_t)
    \,\|\,\pi_0(\cdot\mid s_t)
    \right),
    \label{eq:w2s_objective}
\end{align}
where $\beta>0$ controls the strength of KL regularization.
At a fixed state $s_t$, its gradient is
\begin{align}
    \nabla_\theta \mathcal{J}_{\mathrm{W2S}}
    &=
    \mathbb{E}_{a\sim\pi_\theta}
    \Bigg[
    \Bigg(
    \underbrace{\Delta_t(a)}_{\text{Weak-teacher signal}}
    -
    \underbrace{\beta\log
    \frac{\pi_\theta(a\mid s_t)}
         {\pi_0(a\mid s_t)}}_{\text{KL regularization}}
    \Bigg)
    \nabla_\theta\log\pi_\theta(a\mid s_t)
    \Bigg]. \notag
\end{align}
This objective encourages the student to incorporate the relative improvements encoded by the weak teachers while remaining close to its reference policy.

\paragraph{Closed-Form Contrastive Policy.}

For each fixed prefix, maximizing \Cref{eq:w2s_objective} over the token distribution gives
\begin{align}
    \pi^*(a\mid s_t)
    &=
    \frac{1}{Z(s_t)}
    \pi_0(a\mid s_t)
    \exp\left(
        \frac{\Delta_t(a)}{\beta}
    \right), \notag
\end{align}
where $Z(s_t)=\sum_{a\in\mathcal{V}}\pi_0(a\mid s_t)\exp(\Delta_t(a)/\beta)$ is the normalization constant. Let $\lambda=1/\beta$, and denote the logits of $\pi_0$, $\pi_{T^+}$, and $\pi_{T^-}$ by $z_0$, $z_{T^+}$, and $z_{T^-}$, respectively.
Assuming a shared token vocabulary, the closed-form policy can equivalently be expressed as
\begin{align}
    \pi^*(\cdot\mid s_t)
    &=
    \operatorname{softmax}\left(
        z_0(s_t)
        +\lambda\left[
            z_{T^+}(s_t)-z_{T^-}(s_t)
        \right]
    \right).
    \label{eq:logit_steering}
\end{align}

\Cref{eq:logit_steering} identifies a directly sampleable target for the state-wise objective.
The additive constants relating teacher logits to log probabilities cancel in the softmax.
This is the contrastive logit construction used in decoding-time guidance and W2S-OPD \citep{DExperts,mitchell2024eft,Proxytuning,W2SOPD}.
The state-wise solution motivates evaluating teacher guidance with frozen models; \Cref{sec:theory} characterizes its realizability and optimization in a linear-softmax model, and \Cref{sec:steering_vs_directopd} assesses its correspondence with trained students empirically.
We first examine direct generation and support-aware gating (\Cref{sec:training_free_steering}), then use the resulting trajectories for rejection fine-tuning (\Cref{sec:steering_rft}).

\section{Method}
\label{sec:method}

\subsection{Inference-Time Weak-to-Strong Transfer}
\label{sec:training_free_steering}

We first evaluate the policy in \Cref{eq:logit_steering} with the student and teachers frozen. The reference student $\pi_0$ is Qwen3-4B-Base, and the negative teacher $T^-$ is Qwen3-1.7B-Base. We obtain domain-specific positive teachers $T^+$ by training the 1.7B model with GRPO for up to 400 steps on math, code, and instruction following. The initial comparison uses teacher checkpoints at steps 200 and 400.

\begin{figure*}[t]
    \centering
    \includegraphics[width=\textwidth]{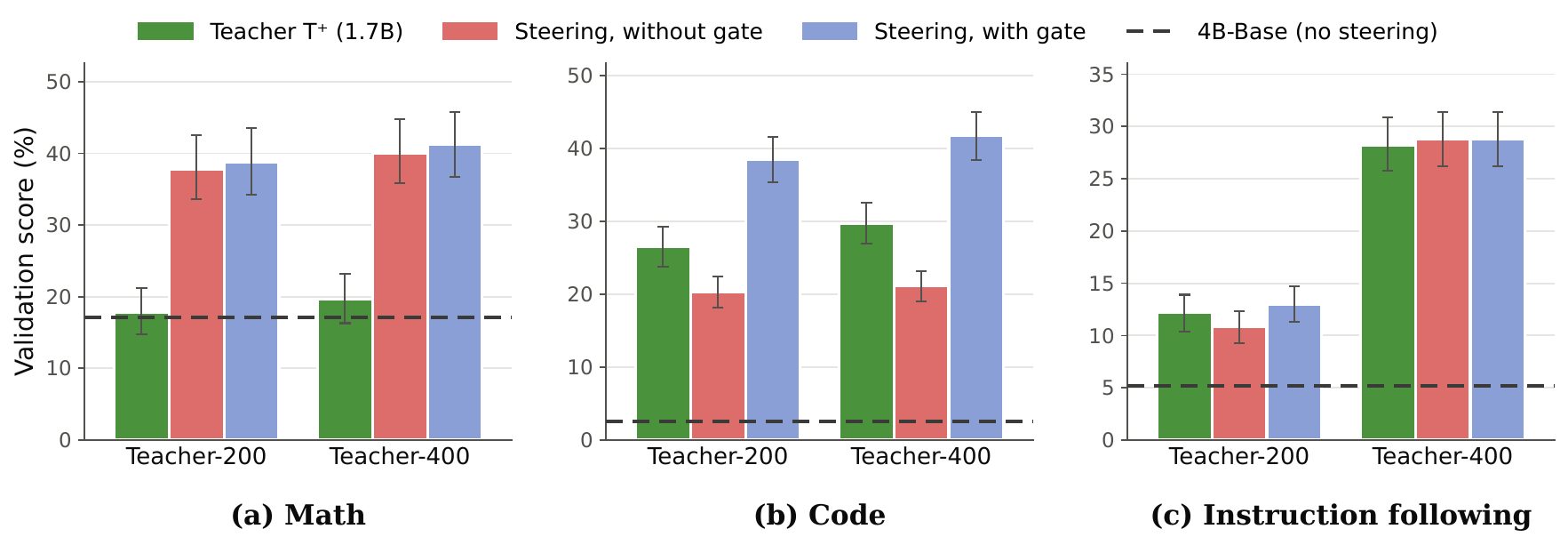}
    \caption{\textbf{Support-aware inference-time transfer.} Validation scores compare the positive teacher, ungated transfer, and gated transfer across domains. Dashed lines mark the reference student.}
    \label{fig:gate_ablation}
\end{figure*}

Direct sampling improves performance in math, where the steered student exceeds both the unsteered student and the positive teacher (\Cref{fig:gate_ablation}). The effect varies across domains: in code, ungated steering remains below the positive teacher. We examine the contrastive signal to improve the reliability of this transfer.

\paragraph{Support-aware steering.}
The steering policy depends on the relative preferences of the weak teachers:
\begin{align}
    \pi^*(a \mid s_t)
    &\propto
    \pi_0(a \mid s_t)
    \exp\left(\lambda \Delta_t(a)\right), \notag
\end{align}
where
\begin{align}
    \Delta_t(a)
    &=
    \log
    \frac{
        \pi_{T^+}(a \mid s_t)
    }{
        \pi_{T^-}(a \mid s_t)
    }. \notag
\end{align}

A token can have a large likelihood ratio even when both teachers assign it negligible probability. For example, probabilities of $10^{-6}$ and $10^{-9}$ give a ratio of $10^3$. The resulting positive increment can promote a token with little support from either teacher.

We introduce a support-aware gate that restricts positive steering increments to tokens supported by at least one teacher. At each decoding step, define
\begin{align}
    \mathcal{S}_{\tau}(s_t)
    &=
    \left\{
        a \in \mathcal{V} :
        \max\left(
            \pi_{T^+}(a \mid s_t),
            \pi_{T^-}(a \mid s_t)
        \right)
        \geq \tau
    \right\}, \notag
\end{align}
where $\tau$ is a probability threshold.

We then modify the steering signal as
\begin{align}
    \widetilde{\Delta}_t(a)
    &=
    \begin{cases}
        \Delta_t(a),
        & a \in \mathcal{S}_{\tau}(s_t),\\
        \min\{\Delta_t(a), 0\},
        & a \notin \mathcal{S}_{\tau}(s_t).
    \end{cases}
    \label{eq:gated_steering_signal}
\end{align}
The resulting gated steering policy is
\begin{align}
    \pi_{\mathrm{gate}}(a \mid s_t)
    &=
    \frac{
        \pi_0(a \mid s_t)
        \exp\left(\lambda\widetilde{\Delta}_t(a)\right)
    }{
        {\textstyle\sum_{u \in \mathcal{V}}}
        \pi_0(u \mid s_t)
        \exp\left(\lambda\widetilde{\Delta}_t(u)\right)
    }.
    \label{eq:gated_steering_policy}
\end{align}

The gate preserves the signal on teacher-supported tokens. Outside this set, it retains negative increments and sets positive increments to zero. The gate acts on the normalized log-probability difference $\Delta_t$, as defined in \Cref{eq:weak_signal}; the gated distribution is obtained by adding $\lambda\widetilde{\Delta}_t$ to the student's logits and applying softmax.

\paragraph{Effectiveness of support-aware gating.}
Gating produces the clearest improvement in code, where it exceeds both ungated steering and the positive teacher (\Cref{fig:gate_ablation}). It also improves math performance. The two variants have similar instruction-following scores, with a modest advantage for gating at teacher step 200. These results establish gated steering as a useful policy for evaluating weak-teacher guidance and generating student responses.

\subsection{Evaluating Teacher Checkpoints before Student Training}
\label{sec:steering_vs_directopd}

We next ask whether evaluating a candidate teacher through steering gives useful information about its downstream Direct-OPD performance. Such a proxy would allow teacher checkpoints to be screened through generation before training the student.

For each domain, we use four positive-teacher checkpoints at GRPO steps 40, 120, 200, and 400. We train a Direct-OPD student for 100 steps with each checkpoint and compare its validation score with that of the gated steering policy using the same teacher pair. All checkpoints use the domain-specific validation sets described in \Cref{app:details}.

\begin{figure*}[t]
    \centering
    \includegraphics[width=\textwidth]{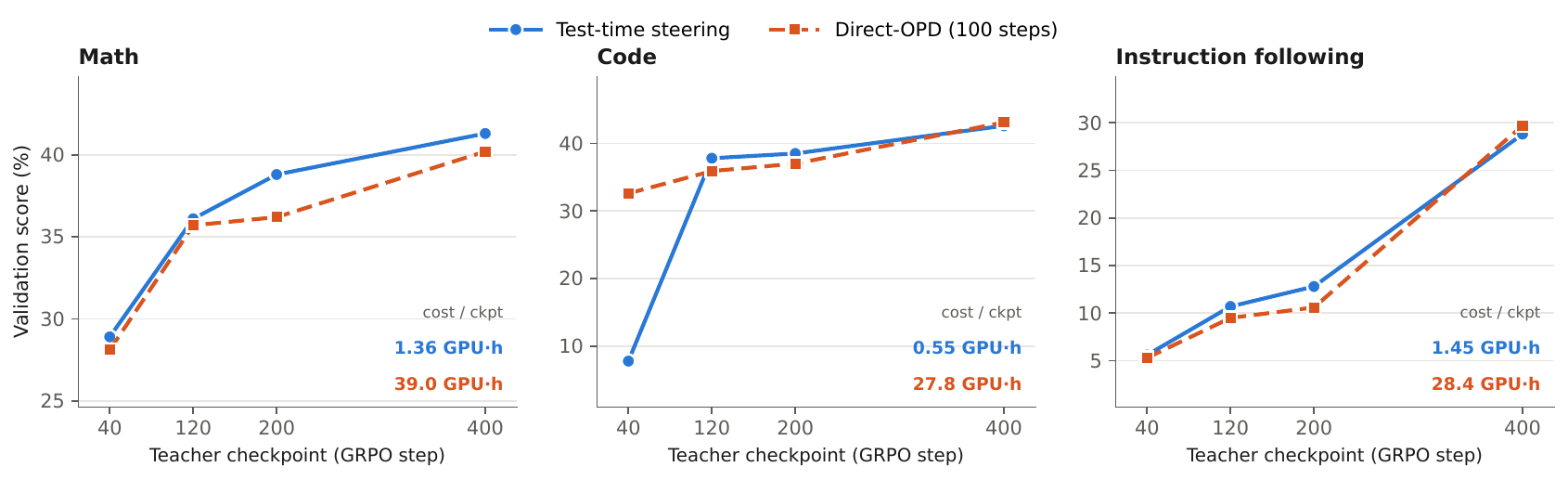}
    \caption{\textbf{Inference-time transfer tracks Direct-OPD across teacher checkpoints.} Validation uses the same teacher pairs; GPU-hours show the mean cost per checkpoint.}
    \label{fig:ckpt_selection}
\end{figure*}

Steering broadly follows the checkpoint trends of Direct-OPD (\Cref{fig:ckpt_selection}). The scores are similar in math and instruction following, and in code at steps 120, 200, and 400. At the step-40 code checkpoint, steering performs substantially below Direct-OPD. This comparison supports using steering to assess candidate guidance, with the correspondence measured across the evaluated checkpoints.

Each steering evaluation uses only forward passes through the frozen models and takes a fraction of the GPU-hours reported for Direct-OPD. Costs in \Cref{fig:ckpt_selection} are averaged over the four teacher checkpoints. This makes additional checkpoints accessible to evaluation before student training. We consider response behavior alongside benchmark scores when selecting guidance: for instruction following, the step-200 teacher is used because later training produces shorter, collapsed responses despite higher scores (\Cref{app:details}).

\subsection{Rejection Fine-Tuning with Inference-Time Guidance}
\label{sec:steering_rft}

Steering provides both a way to evaluate teacher guidance and a policy for generating student-conditioned responses. Its per-token computation includes the student and both teachers. In our vLLM implementation on a single RTX Pro 6000, throughput is 2.4k tokens/s with steering and 5.6k tokens/s for the student alone, measured with 1k-token generations.

We use rejection fine-tuning (RFT) \citep{rft} to incorporate the steering-generated behavior into the student's parameters. \methodname{} samples $K$ trajectories per training prompt from $\pi_{\mathrm{gate}}$ (\Cref{eq:gated_steering_policy}), retains nondegenerate responses that pass task-specific verification and format checks, and applies supervised fine-tuning (SFT) to the accepted responses. \Cref{alg:steering_rft} describes the procedure. For each prompt, the retained responses follow the steering distribution conditioned on acceptance. Maximum-likelihood fitting therefore targets the accepted response distribution, with the retained sample counts determining the relative contribution of each prompt. In the main experiments, $K=1$.

For mixed-domain training, we generate responses using each prompt's domain-specific positive teacher and the shared negative teacher, then pool the accepted trajectories for SFT. The trained student generates independently of either teacher. The same weak-teacher pair can also be reused with a different reference student by generating new student-conditioned trajectories.

\begin{algorithm}[t]
    \caption{\methodname{}}
    \label{alg:steering_rft}
    \small
    \noindent\textbf{Input:} reference student $\pi_0$; weak teachers $\pi_{T^+}$, $\pi_{T^-}$; prompts $\mathcal{D}$; verifier $R$; steering strength $\lambda$; support threshold $\tau$; samples per prompt $K$
    \begin{enumerate}[label=\arabic*:,leftmargin=2.1em,itemsep=0.12em,topsep=0.5em,parsep=0pt]
        \item Initialize an empty dataset $\mathcal{B}$.
        \item \textbf{for} each prompt $x\in\mathcal{D}$ \textbf{do}
        \item \hspace*{1em}\textbf{for} $k=1,\ldots,K$ \textbf{do}
        \item \hspace*{2em}Sample $y\sim\pi_{\mathrm{gate}}(\cdot\mid x)$ token by token using \Cref{eq:gated_steering_signal,eq:gated_steering_policy}.
        \item \hspace*{2em}\textbf{if} $R(x,y)=1$ and $y$ passes the format checks and is not degenerate \textbf{then}
        \item \hspace*{3em}Append $(x,y)$ to $\mathcal{B}$.
        \item \hspace*{2em}\textbf{end if}
        \item \hspace*{1em}\textbf{end for}
        \item \textbf{end for}
        \item Initialize $\pi_\theta\gets\pi_0$ and fine-tune $\pi_\theta$ on $\mathcal{B}$ with SFT.
        \item \textbf{return} $\pi_\theta$
    \end{enumerate}
\end{algorithm}

\Cref{sec:exp} evaluates this procedure in single-domain and mixed-domain settings, compares data generated by different rollout policies, and examines reuse of the teachers across student scales.

\section{Theoretical Justification}
\label{sec:theory}

Notably, the 
To further understand why the weak-to-strong OPD helps in improving the sample efficiency and how

We study when the closed-form policy in \Cref{eq:logit_steering} belongs to the student's model family and can be reached by OPD optimization.

\paragraph{Linear-softmax preliminaries.}
Fix prefixes $s_1,\ldots,s_n$ with weights $\mu_i>0$, $\sum_i\mu_i=1$, and a shared vocabulary of size $V$.
Let $\mathbf{A}\in\mathbb{R}^{n\times d_S}$ and $\mathbf{B}\in\mathbb{R}^{n\times d_T}$ collect fixed student and teacher features, with rows $\boldsymbol{a}_i^\top$ and $\boldsymbol{b}_i^\top$.
The student head $\mathbf{W}\in\mathbb{R}^{d_S\times V}$ and teacher heads $\mathbf{\Theta}_\pm\in\mathbb{R}^{d_T\times V}$ define $\boldsymbol{p}_i(\mathbf{W})=\operatorname{softmax}(\mathbf{W}^\top \boldsymbol{a}_i)$ and $\boldsymbol{p}_i^\pm=\operatorname{softmax}(\mathbf{\Theta}_\pm^\top \boldsymbol{b}_i)$.
Write $\mathbf{W}_0$ for the reference student head and let $\mathbf{\Delta}$ stack the signals in \Cref{eq:weak_signal}.
Centering across tokens removes softmax normalization constants:
\begin{align}
    \mathbf{C}&=\mathbf{I}_V-\frac{\boldsymbol{1}\boldsymbol{1}^\top}{V},
    \qquad \mathbf{D}=\mathbf{\Delta} \mathbf{C}=\mathbf{B}(\mathbf{\Theta}_+-\mathbf{\Theta}_-)\mathbf{C}. \notag
\end{align}
Thus the inference-time target is $\boldsymbol{q}_i=\operatorname{softmax}(\mathbf{W}_0^\top \boldsymbol{a}_i+\mathbf{D}_{i,:}^\top/\beta)$.
Maximizing the $\boldsymbol{\mu}$-weighted objective in \Cref{eq:w2s_objective} is equivalent to minimizing
\begin{align}
    \mathcal{L}(\mathbf{W})&=\beta\sum_{i=1}^n\mu_iD_{\mathrm{KL}}\bigl(\boldsymbol{p}_i(\mathbf{W})\,\|\,\boldsymbol{q}_i\bigr).
    \label{eq:theory_loss}
\end{align}
Let $\mathbf{\Sigma}=\mathbf{A}^\top\operatorname{diag}(\boldsymbol{\mu})\mathbf{A}$ and $\Lambda=\lambda_{\max}(\mathbf{\Sigma})>0$.

\begin{theorem}[Realizability and tracking of the closed form]
\label{thm:opd_tracking}
A student head realizes $\boldsymbol{q}_i$ at every prefix if and only if
$(\mathbf{I}-\mathbf{A}\mathbf{A}^\dagger)\mathbf{D}=0$, where $\mathbf{A}^\dagger$ is the Moore--Penrose pseudoinverse.
Under this coverage condition,
\begin{align}
    \mathbf{W}^\star&=\mathbf{W}_0+\beta^{-1}\mathbf{A}^\dagger \mathbf{D},
    \qquad \boldsymbol{p}_i(\mathbf{W}^\star)=\boldsymbol{q}_i. \notag
\end{align}
Gradient descent $\mathbf{W}_{k+1}=\mathbf{W}_k-\eta\nabla\mathcal{L}(\mathbf{W}_k)$, initialized at $\mathbf{W}_0$, satisfies
\begin{align}
    \|\mathbf{W}_k-\mathbf{W}^\star\|_F&\le\rho^k\|\mathbf{W}_0-\mathbf{W}^\star\|_F \notag
\end{align}
for some $\rho\in[0,1)$ whenever $0<\eta\le(\beta\Lambda)^{-1}$.
\end{theorem}

\begin{remark}[Coverage and efficient weak-to-strong transfer]
The teacher change has $\operatorname{rank}(\mathbf{D})\le\operatorname{rank}(\mathbf{B})$.
When richer student features cover these changed directions, the closed form directly evaluates a realizable student policy through frozen-model forward passes.
Coverage is required only for the change: components shared by the two teachers cancel, so the student can represent their difference even when the full teacher policy lies outside its feature span.
This enables candidate guidance to be evaluated without a separate student optimization run.
\end{remark}

\begin{remark}[Trackability and optimization]
The closed form is also the limit of ordinary head gradient descent, linking inference-time evaluation to the policy approached by conditional OPD optimization.
The proof factors the exact gradient into a softmax Fisher operator acting on one shared head error; its rate depends on the positive feature spectrum and token probabilities.
At the solution, the loss is locally strongly convex on the identifiable head subspace (\Cref{prop:tracking_local_curvature}).
This local geometry supports stochastic-gradient analyses and fixed positive-definite preconditioned updates under appropriate step-size, noise, and stability assumptions \citep{bottou2018optimization,uschmajew2022note}.
\end{remark}

\Cref{app:tracking} gives the explicit rate, supporting lemmas, and complete proof, together with an approximation bound for partial coverage and support-gated guidance.

\section{Experiments}
\label{sec:exp}

\subsection{Experimental Setup}
\label{sec:exp_setup}

\paragraph{Datasets and models.}
We evaluate \methodname{} on math, code, and instruction following (IF), using training prompts from DAPO-Math-17k, code-r1-12k, and IF-Nemotron, respectively \citep{dapo,code-r1,nemo-gym}. Unless otherwise specified, the student is initialized from Qwen3-4B-Base \citep{qwen3}, which also defines the reference policy $\pi_0$, and the negative teacher $T^-$ is Qwen3-1.7B-Base. We obtain each domain's positive teacher $T^+$ by training Qwen3-1.7B-Base with GRPO. We use the step-400 checkpoints for math and code and the step-200 checkpoint for IF. The IF teacher develops short, low-entropy responses with further training, motivating the earlier checkpoint (\Cref{app:details}).

\paragraph{Training and sampling budgets.}
For the W2S-OPD baselines, we follow the Direct-OPD training protocol \citep{DirectOPD}. Single-domain runs use 100 steps, 64 prompts per step, and four responses per prompt, totaling 6,400 prompt instances per domain. Mixed-domain runs use 300 steps and interleave 6,400 prompt instances from each domain, totaling 19,200. \methodname{} uses the same prompt-instance budgets, generates one trajectory per prompt instance, and applies rejection fine-tuning to the verified trajectories.

\paragraph{Evaluation.}
The steering analyses in \Cref{fig:gate_ablation,fig:ckpt_selection} use separate domain-specific validation sets. Their construction and evaluation settings are detailed in \Cref{app:details}.

For the main experiments, we evaluate math on AIME 2024, AIME 2025, and AMC 2023 \citep{aime24,aime25,maa_amc}, using 16 samples per question (avg@16) and a maximum generation length of 8,192 tokens. Math denotes the unweighted mean accuracy across these three benchmarks. We additionally evaluate on MATH500, Minerva, and OlympiadBench \citep{MATH500,minerva,olympiabench} using avg@4 with the same token limit, reporting the mean across all six benchmarks in \Cref{tab:math_generalization}. Code denotes pooled avg@4 across HumanEval+ and MBPP+ \citep{humaneval}, with a maximum generation length of 4,096 tokens. For IF, we report strict prompt-level accuracy on IFEval \citep{IFeval}, averaged over four generations per prompt with the same token limit. In mixed-domain comparisons, Avg. is the unweighted mean of Math, Code, and IF. Additional implementation details are provided in \Cref{app:details}.

\subsection{Main Results}
\label{sec:main_results}

\begin{figure*}[t]
    \centering
    \includegraphics[width=\textwidth]{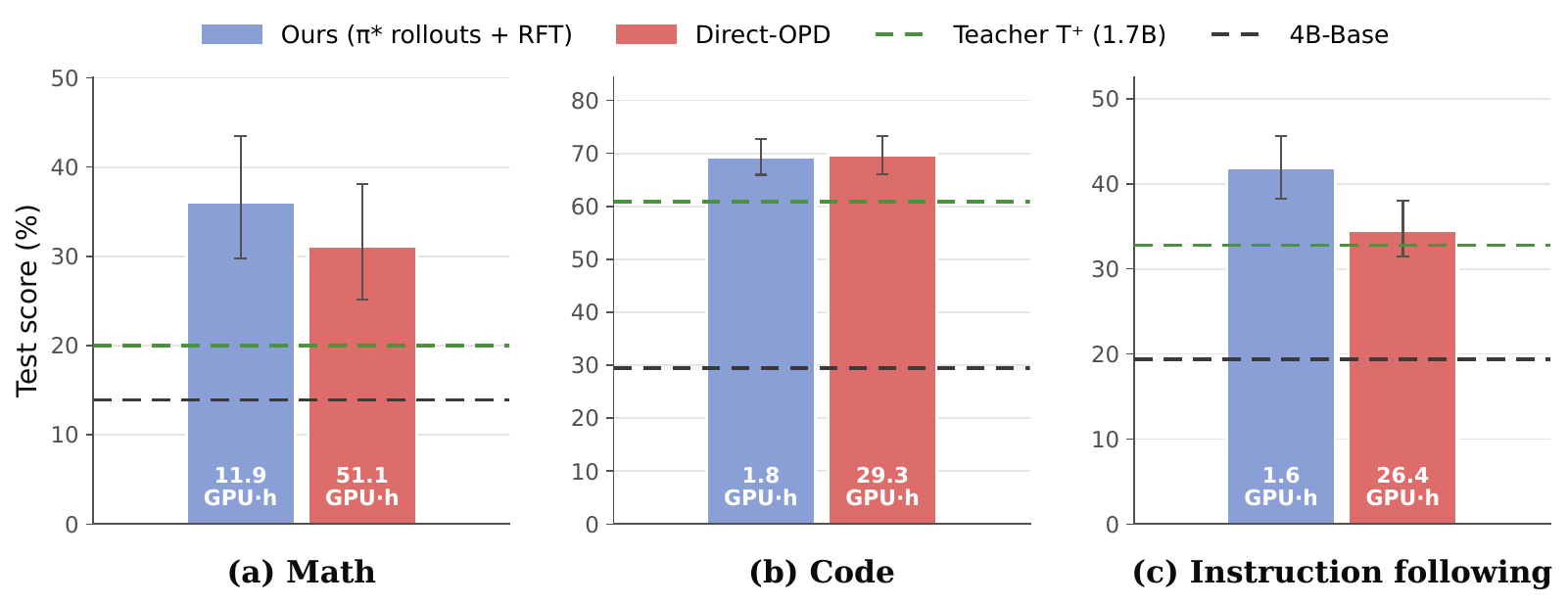}
    \caption{\textbf{Single-domain performance and cost.} \methodname{} and Direct-OPD use matched prompt budgets. GPU-hours include rollout generation and student training.}
    \label{fig:main_results}
\end{figure*}

\paragraph{Single-domain comparison with OPD.}
We first train a separate student for each domain and compare \methodname{} with Direct-OPD. \Cref{fig:main_results} shows comparable or higher performance for \methodname{} with lower computational cost for rollout generation and student training. These results show that verified steering trajectories provide effective supervision for transferring weak-teacher guidance into the student.

Across six mathematical reasoning benchmarks, \methodname{} exceeds Direct-OPD and Waypoint-OPD \citep{waypointOPD} on five benchmarks (\Cref{tab:math_generalization}). Its mean accuracy is 44.2\%, compared with 41.4\% for Direct-OPD and 42.3\% for Waypoint-OPD. The gains extend across competition mathematics, MATH500, and OlympiadBench, while both OPD baselines score higher on Minerva.

\begin{table*}[t]
    \centering
    \small
    \setlength{\tabcolsep}{6pt}
    \caption{
        \textbf{Mathematical reasoning performance across six benchmarks.}
        Comparison with Direct-OPD and Waypoint-OPD, both trained
        for 100 steps.
        AIME24, AIME25, and AMC23 use avg@16, while
        MATH500, Minerva, and OlympiadBench use avg@4.
        All evaluations use a maximum generation length of
        8,192 tokens.
        Avg. denotes the unweighted mean across benchmarks.
    }
    \label{tab:math_generalization}
    \begin{adjustbox}{max width=\textwidth}
    \begin{tabular}{lccccccc}
        \toprule
        \textbf{Method}
        & \textbf{AIME24}
        & \textbf{AIME25}
        & \textbf{AMC23}
        & \textbf{MATH500}
        & \textbf{Minerva}
        & \textbf{OlympiadBench}
        & \textbf{Avg.} \\
        \midrule
        Direct-OPD
        & 19.8 & 17.1 & 56.4 & 80.5
        & \textbf{30.1} & 44.6 & 41.4 \\
        Waypoint-OPD
        & 19.6 & 19.2 & 60.2 & 79.8
        & \textbf{30.1} & 44.8 & 42.3 \\
        \midrule
        \methodname{}
        & \textbf{24.2} & \textbf{19.6}
        & \textbf{64.5} & \textbf{82.0}
        & 28.7 & \textbf{46.3} & \textbf{44.2} \\
        \bottomrule
    \end{tabular}
    \end{adjustbox}
\end{table*}

\paragraph{Mixed-domain comparison with OPD.}
We next combine the three domains within a single student and compare \methodname{} with Routed Direct-OPD and $\Delta$-MOPD \citep{delta-mopd}. Under the matched prompt-instance budgets, \methodname{} achieves an average score of 47.1\%, compared with 44.7\% for Routed Direct-OPD and 45.5\% for $\Delta$-MOPD (\Cref{tab:main_efficiency}). It scores highest on math and IF, while $\Delta$-MOPD scores highest on code. The recorded cost is 14.8 GPU-hours for \methodname{}, compared with 108.4 and 100.4 GPU-hours for the two OPD baselines. Mixing verified steering trajectories therefore supports transfer across domains in one RFT run.

\begin{table*}[t]
    \centering
    \small
    \setlength{\tabcolsep}{6pt}
    \caption{
        \textbf{Performance and computational efficiency in the mixed-domain setting.}
        Comparison with Routed Direct-OPD and $\Delta$-MOPD
        under matched prompt budgets.
        Both OPD baselines are trained for 300 steps.
        Results for \methodname{} are averaged over three random seeds.
        Avg. denotes the unweighted mean across the three domains.
        GPU-hours report the measured computational cost,
        and Cost Ratio is normalized by the cost of \methodname{}.
        Bold indicates the best performance or lowest cost.
    }
    \label{tab:main_efficiency}
    \begin{adjustbox}{max width=\textwidth}
    \begin{tabular}{lcccccc}
        \toprule
        \textbf{Method}
        & \textbf{Math}
        & \textbf{Code}
        & \textbf{IFEval}
        & \textbf{Avg.}
        & \textbf{GPU-hours}
        & \textbf{Cost Ratio} \\
        \midrule
        \methodname{}
        & \textbf{33.9} & 68.3 & \textbf{39.1}
        & \textbf{47.1} & \textbf{14.8} & \textbf{1.0$\times$} \\
        \midrule
        $\Delta$-MOPD
        & 31.0 & \textbf{69.7} & 35.8
        & 45.5 & 100.4 & 6.8$\times$ \\
        Routed Direct-OPD
        & 31.8 & 67.6 & 34.8
        & 44.7 & 108.4 & 7.3$\times$ \\
        \bottomrule
    \end{tabular}
    \end{adjustbox}
\end{table*}

\paragraph{Comparison with stronger teachers.}
We compare steering-generated data with math trajectories sampled from Qwen3-14B, Qwen3-8B, and Gemma-3-12B-IT \citep{gemmateam2025gemma3technicalreport}. Qwen3-14B and Qwen3-8B use non-thinking mode. All methods use the same single-domain prompt budget and RFT procedure, retaining trajectories with correct final answers.

Steering produces fewer correct training trajectories than the stronger teachers but yields higher accuracy after RFT on the 4B student (\Cref{tab:strong_teacher_rft}). At four epochs, steering-generated data yields 36.1\% accuracy, compared with 33.1\% for Qwen3-14B data, with 2,623 retained trajectories compared with 3,467. Across all reported epochs, the highest Qwen3-14B-data result is 33.9\%. These results show that source-model accuracy and the number of correct trajectories alone do not determine downstream RFT performance. Steering incorporates the student's distribution at every decoding step, making student-conditioned generation a useful direction for constructing transfer data.

\paragraph{Transfer across student scales.}
We apply the same 1.7B teacher pair to Qwen3-14B-Base without additional teacher training. At the default four epochs, \methodname{} achieves 41.6\% accuracy, compared with 32.6\% for self-generated RFT (\Cref{tab:strong_teacher_rft}). Across the reported 2-, 4-, 6-, and 8-epoch checkpoints, the respective peak accuracies are 43.4\% and 32.7\%. Direct-OPD on the 14B student exceeds our available compute budget and is not included in this comparison.

The results show that one weak-teacher pair can guide both 4B and 14B students. Each student uses its own steered trajectories for RFT, while reusing the teacher training.

\begin{table*}[t]
    \centering
    \small
    \setlength{\tabcolsep}{6pt}
    \caption{
        \textbf{Comparison with stronger teachers and transfer across student scales on math.}
        All methods follow the same single-domain prompt budget,
        retaining only trajectories with correct final answers for RFT.
        Qwen3-8B and Qwen3-14B operate in non-thinking mode
        when serving as stronger rollout teachers.
        Source Perf. denotes the average accuracy of the rollout
        policy on AIME 2024, AIME 2025, and AMC 2023 (avg@16).
        \# Correct denotes the number of retained training trajectories.
        The last four columns report the same math evaluation metric
        after 2, 4, 6, and 8 epochs of RFT.
        Bold indicates the best RFT result within each student scale.
    }
    \label{tab:strong_teacher_rft}
    \begin{adjustbox}{max width=\textwidth}
    \begin{tabular}{llcccccc}
        \toprule
        \textbf{Student} &
        \textbf{Rollout Policy} &
        \textbf{Source Perf.} &
        \textbf{\# Correct} &
        \multicolumn{4}{c}{\textbf{RFT Epochs}} \\
        \cmidrule(lr){5-8}
        & & \textbf{(\%)} & &
        \textbf{2} & \textbf{4} & \textbf{6} & \textbf{8} \\
        \midrule
        \multirow{5}{*}{4B-Base}
        & $\pi_{\mathrm{gate}}$ (Ours)
        & 33.0 & 2,623 & 32.9 & \textbf{36.1} & 34.5 & 35.7 \\
        & Qwen3-14B
        & 43.3 & 3,467 & 30.5 & 33.1 & 33.9 & 31.2 \\
        & Qwen3-8B
        & 38.9 & 3,136 & 28.6 & 30.9 & 30.7 & 28.8 \\
        & Gemma-3-12B-IT
        & 37.3 & 2,961 & 26.0 & 25.9 & 25.3 & 25.3 \\
        & Self (4B-Base)
        & 13.6 & 209 & 12.7 & 20.8 & 17.9 & 24.8 \\
        \midrule
        \multirow{2}{*}{14B-Base}
        & $\pi_{\mathrm{gate}}$ (Ours)
        & 41.1 & 3,259 & 41.8 & 41.6 & \textbf{43.4} & 41.5 \\
        & Self (14B-Base)
        & 25.9 & 922 & 28.2 & 32.6 & 30.2 & 32.7 \\
        \bottomrule
    \end{tabular}
    \end{adjustbox}
\end{table*}

\section{Conclusion}
\label{sec:conclusion}

Inference-time weak-to-strong transfer makes teacher guidance directly evaluable before training a stronger student. Support-aware steering broadly follows Direct-OPD performance across teacher checkpoints and domains, making it useful for evaluating candidate teachers. \methodname{} turns verified steering trajectories into supervision for rejection fine-tuning and achieves comparable or higher performance than the evaluated OPD baselines at lower computational cost. The same weak-teacher pair transfers across student sizes, and steering-generated data yields stronger RFT results than data sampled from more capable models in our math experiments.

\begin{itemize}[leftmargin=*]
    \item \textbf{Coverage yields efficient and trackable transfer.}
    Our linear-softmax analysis shows that covering the teacher's centered logit change makes the closed-form policy exactly realizable and the limit of ordinary head gradient descent.
    This connects frozen-model evaluation to conditional OPD optimization and explains how weak-teacher guidance can be assessed within a richer student without retraining it for each candidate.
\end{itemize}

\bibliography{references}

\begin{thebibliography}{37}
\providecommand{\natexlab}[1]{#1}
\providecommand{\url}[1]{\texttt{#1}}
\expandafter\ifx\csname urlstyle\endcsname\relax
  \providecommand{\doi}[1]{doi: #1}\else
  \providecommand{\doi}{doi: \begingroup \urlstyle{rm}\Url}\fi

\bibitem[Agarwal et~al.(2024)Agarwal, Vieillard, Zhou, Stanczyk, Garea, Geist,
  and Bachem]{gkd}
R.~Agarwal, N.~Vieillard, Y.~Zhou, P.~Stanczyk, S.~R. Garea, M.~Geist, and
  O.~Bachem.
\newblock On-policy distillation of language models: Learning from
  self-generated mistakes.
\newblock In \emph{The Twelfth International Conference on Learning
  Representations}, 2024.

\bibitem[Anonymous(2026)]{delta-mopd}
Anonymous.
\newblock Delta-{MOPD}: Cross-tokenizer multi-teacher on-policy distillation.
\newblock In \emph{Submitted to The Fifteenth International Conference on
  Learning Representations}, 2026.
\newblock under review.

\bibitem[Bottou et~al.(2018)Bottou, Curtis, and
  Nocedal]{bottou2018optimization}
L.~Bottou, F.~E. Curtis, and J.~Nocedal.
\newblock Optimization methods for large-scale machine learning.
\newblock \emph{SIAM Review}, 60\penalty0 (2):\penalty0 223--311, 2018.
\newblock \doi{10.1137/16M1080173}.

\bibitem[Feng et~al.(2026)Feng, ang Gao, Chi, Wu, Zhang, Jiang, He, Ma, Zhang,
  and Zhou]{DirectOPD}
S.~Feng, H.~ang Gao, H.~Chi, H.~Wu, Z.~Zhang, Z.~Jiang, B.~He, W.-Y. Ma, Y.-Q.
  Zhang, and H.~Zhou.
\newblock Weak-to-strong generalization via direct on-policy distillation,
  2026.

\bibitem[Gulcehre et~al.(2023)Gulcehre, Paine, Srinivasan, Konyushkova, Weerts,
  Sharma, Siddhant, Ahern, Wang, Gu, Macherey, Doucet, Firat, and
  de~Freitas]{rest}
C.~Gulcehre, T.~L. Paine, S.~Srinivasan, K.~Konyushkova, L.~Weerts, A.~Sharma,
  A.~Siddhant, A.~Ahern, M.~Wang, C.~Gu, W.~Macherey, A.~Doucet, O.~Firat, and
  N.~de~Freitas.
\newblock Reinforced self-training ({ReST}) for language modeling, 2023.

\bibitem[He et~al.(2024)He, Luo, Bai, Hu, Thai, Shen, Hu, Han, Huang, Zhang,
  Liu, Qi, Liu, and Sun]{olympiabench}
C.~He, R.~Luo, Y.~Bai, S.~Hu, Z.~L. Thai, J.~Shen, J.~Hu, X.~Han, Y.~Huang,
  Y.~Zhang, J.~Liu, L.~Qi, Z.~Liu, and M.~Sun.
\newblock Olympiadbench: A challenging benchmark for promoting agi with
  olympiad-level bilingual multimodal scientific problems, 2024.

\bibitem[Hendrycks et~al.(2021)Hendrycks, Burns, Kadavath, Arora, Basart, Tang,
  Song, and Steinhardt]{MATH500}
D.~Hendrycks, C.~Burns, S.~Kadavath, A.~Arora, S.~Basart, E.~Tang, D.~Song, and
  J.~Steinhardt.
\newblock Measuring mathematical problem solving with the math dataset.
\newblock \emph{NeurIPS}, 2021.

\bibitem[Hou et~al.(2026)Hou, Jiang, Quan, Li, Feng, Liu, and
  Zhang]{waypointOPD}
B.~Hou, G.~Jiang, G.~Quan, W.~Li, W.~Feng, G.~Liu, and Y.~Zhang.
\newblock Waypoint-opd: Transferring rl-induced policy shifts via gated
  on-policy distillation, 2026.

\bibitem[Ichinose et~al.(2026)Ichinose, Ma, Oi, Koike, and Okazaki]{CODIT}
T.~Ichinose, Y.~Ma, M.~Oi, R.~Koike, and N.~Okazaki.
\newblock Synthesizing instruction-tuning datasets with contrastive decoding,
  2026.

\bibitem[Lewkowycz et~al.(2022)Lewkowycz, Andreassen, Dohan, Dyer, Michalewski,
  Ramasesh, Slone, Anil, Schlag, Gutman-Solo, Wu, Neyshabur, Gur-Ari, and
  Misra]{minerva}
A.~Lewkowycz, A.~Andreassen, D.~Dohan, E.~Dyer, H.~Michalewski, V.~Ramasesh,
  A.~Slone, C.~Anil, I.~Schlag, T.~Gutman-Solo, Y.~Wu, B.~Neyshabur,
  G.~Gur-Ari, and V.~Misra.
\newblock Solving quantitative reasoning problems with language models.
\newblock In S.~Koyejo, S.~Mohamed, A.~Agarwal, D.~Belgrave, K.~Cho, and A.~Oh,
  editors, \emph{Advances in Neural Information Processing Systems}, volume~35,
  pages 3843--3857. Curran Associates, Inc., 2022.
\newblock \doi{10.52202/068431-0278}.

\bibitem[Li et~al.(2023)Li, Holtzman, Fried, Liang, Eisner, Hashimoto,
  Zettlemoyer, and Lewis]{ContrastiveDecoding}
X.~L. Li, A.~Holtzman, D.~Fried, P.~Liang, J.~Eisner, T.~Hashimoto,
  L.~Zettlemoyer, and M.~Lewis.
\newblock Contrastive decoding: Open-ended text generation as optimization.
\newblock In A.~Rogers, J.~Boyd-Graber, and N.~Okazaki, editors,
  \emph{Proceedings of the 61st Annual Meeting of the {A}ssociation for
  {C}omputational {L}inguistics (Volume 1: Long Papers)}, pages 12286--12312,
  Toronto, Canada, July 2023. Association for Computational Linguistics.
\newblock \doi{10.18653/v1/2023.acl-long.687}.

\bibitem[Liu et~al.(2021)Liu, Sap, Lu, Swayamdipta, Bhagavatula, Smith, and
  Choi]{DExperts}
A.~Liu, M.~Sap, X.~Lu, S.~Swayamdipta, C.~Bhagavatula, N.~A. Smith, and
  Y.~Choi.
\newblock {DE}xperts: Decoding-time controlled text generation with experts and
  anti-experts.
\newblock In C.~Zong, F.~Xia, W.~Li, and R.~Navigli, editors, \emph{Proceedings
  of the 59th Annual Meeting of the {A}ssociation for {C}omputational
  {L}inguistics and the 11th International Joint Conference on Natural Language
  Processing (Volume 1: Long Papers)}, pages 6691--6706, Online, Aug. 2021.
  Association for Computational Linguistics.
\newblock \doi{10.18653/v1/2021.acl-long.522}.

\bibitem[Liu et~al.(2024)Liu, Han, Wang, Tsvetkov, Choi, and
  Smith]{Proxytuning}
A.~Liu, X.~Han, Y.~Wang, Y.~Tsvetkov, Y.~Choi, and N.~A. Smith.
\newblock Tuning language models by proxy.
\newblock In \emph{First Conference on Language Modeling}, 2024.

\bibitem[Liu and Zhang(2025)]{code-r1}
J.~Liu and L.~Zhang.
\newblock Code-r1: Reproducing r1 for code with reliable rewards.
\newblock 2025.

\bibitem[Liu et~al.(2023)Liu, Xia, Wang, and ZHANG]{humaneval}
J.~Liu, C.~S. Xia, Y.~Wang, and L.~ZHANG.
\newblock Is your code generated by chatgpt really correct? rigorous evaluation
  of large language models for code generation.
\newblock In A.~Oh, T.~Naumann, A.~Globerson, K.~Saenko, M.~Hardt, and
  S.~Levine, editors, \emph{Advances in Neural Information Processing Systems},
  volume~36, pages 21558--21572. Curran Associates, Inc., 2023.
\newblock \doi{10.52202/075280-0943}.

\bibitem[Lu and Lab(2025)]{opd2}
K.~Lu and T.~M. Lab.
\newblock On-policy distillation.
\newblock \emph{Thinking Machines Lab: Connectionism}, 2025.
\newblock \doi{10.64434/tml.20251026}.
\newblock https://thinkingmachines.ai/blog/on-policy-distillation.

\bibitem[{Mathematical Association of America}(2025)]{maa_amc}
{Mathematical Association of America}.
\newblock American mathematics competitions, 2025.
\newblock Accessed: 2025-09-13.

\bibitem[Mitchell et~al.(2024)Mitchell, Rafailov, Sharma, Finn, and
  Manning]{mitchell2024eft}
E.~Mitchell, R.~Rafailov, A.~Sharma, C.~Finn, and C.~Manning.
\newblock An emulator for fine-tuning large language models using small
  language models.
\newblock In B.~Kim, Y.~Yue, S.~Chaudhuri, K.~Fragkiadaki, M.~Khan, and Y.~Sun,
  editors, \emph{International Conference on Learning Representations}, volume
  2024, pages 13229--13244, 2024.

\bibitem[Muennighoff et~al.(2025)Muennighoff, Yang, Shi, Li, Fei-Fei,
  Hajishirzi, Zettlemoyer, Liang, Cand{\`e}s, and Hashimoto]{s1}
N.~Muennighoff, Z.~Yang, W.~Shi, X.~L. Li, L.~Fei-Fei, H.~Hajishirzi,
  L.~Zettlemoyer, P.~Liang, E.~Cand{\`e}s, and T.~Hashimoto.
\newblock {s1}: Simple test-time scaling.
\newblock In \emph{Proceedings of the 2025 Conference on Empirical Methods in
  Natural Language Processing}, pages 20275--20321. Association for
  Computational Linguistics, Nov. 2025.
\newblock \doi{10.18653/v1/2025.emnlp-main.1025}.

\bibitem[NVIDIA(2025)]{nemo-gym}
NVIDIA.
\newblock Nemo gym: An open source library for scaling reinforcement learning
  environments for llm.
\newblock \url{https://github.com/NVIDIA-NeMo/Gym}, 2025.
\newblock GitHub repository.

\bibitem[Snell et~al.(2025)Snell, Lee, Xu, and Kumar]{snell2025scaling}
C.~Snell, J.~Lee, K.~Xu, and A.~Kumar.
\newblock Scaling {LLM} test-time compute optimally can be more effective than
  scaling parameters for reasoning.
\newblock In \emph{The Thirteenth International Conference on Learning
  Representations}, 2025.

\bibitem[Team et~al.(2025)Team, Kamath, Ferret, Pathak, Vieillard, Merhej,
  Perrin, Matejovicova, Ramé, Rivière, Rouillard, Mesnard, Cideron, bastien
  Grill, Ramos, Yvinec, Casbon, Pot, Penchev, Liu, Visin, Kenealy, Beyer, Zhai,
  Tsitsulin, Busa-Fekete, Feng, Sachdeva, Coleman, Gao, Mustafa, Barr,
  Parisotto, Tian, Eyal, Cherry, Peter, Sinopalnikov, Bhupatiraju, Agarwal,
  Kazemi, Malkin, Kumar, Vilar, Brusilovsky, Luo, Steiner, Friesen, Sharma,
  Sharma, Gilady, Goedeckemeyer, Saade, Feng, Kolesnikov, Bendebury, Abdagic,
  Vadi, György, Pinto, Das, Bapna, Miech, Yang, Paterson, Shenoy, Chakrabarti,
  Piot, Wu, Shahriari, Petrini, Chen, Lan, Choquette-Choo, Carey, Brick,
  Deutsch, Eisenbud, Cattle, Cheng, Paparas, Sreepathihalli, Reid, Tran, Zelle,
  Noland, Huizenga, Kharitonov, Liu, Amirkhanyan, Cameron, Hashemi,
  Klimczak-Plucińska, Singh, Mehta, Lehri, Hazimeh, Ballantyne, Szpektor,
  Nardini, Pouget-Abadie, Chan, Stanton, Wieting, Lai, Orbay, Fernandez,
  Newlan, yeong Ji, Singh, Black, Yu, Hui, Vodrahalli, Greff, Qiu, Valentine,
  Coelho, Ritter, Hoffman, Watson, Chaturvedi, Moynihan, Ma, Babar, Noy, Byrd,
  Roy, Momchev, Chauhan, Sachdeva, Bunyan, Botarda, Caron, Rubenstein,
  Culliton, Schmid, Sessa, Xu, Stanczyk, Tafti, Shivanna, Wu, Pan, Rokni,
  Willoughby, Vallu, Mullins, Jerome, Smoot, Girgin, Iqbal, Reddy, Sheth,
  Põder, Bhatnagar, Panyam, Eiger, Zhang, Liu, Yacovone, Liechty, Kalra, Evci,
  Misra, Roseberry, Feinberg, Kolesnikov, Han, Kwon, Chen, Chow, Zhu, Wei,
  Egyed, Cotruta, Giang, Kirk, Rao, Black, Babar, Lo, Moreira, Martins,
  Sanseviero, Gonzalez, Gleicher, Warkentin, Mirrokni, Senter, Collins, Barral,
  Ghahramani, Hadsell, Matias, Sculley, Petrov, Fiedel, Shazeer, Vinyals, Dean,
  Hassabis, Kavukcuoglu, Farabet, Buchatskaya, Alayrac, Anil, Dmitry, Lepikhin,
  Borgeaud, Bachem, Joulin, Andreev, Hardin, Dadashi, and
  Hussenot]{gemmateam2025gemma3technicalreport}
G.~Team, A.~Kamath, J.~Ferret, S.~Pathak, N.~Vieillard, R.~Merhej, S.~Perrin,
  T.~Matejovicova, A.~Ramé, M.~Rivière, L.~Rouillard, T.~Mesnard, G.~Cideron,
  J.~bastien Grill, S.~Ramos, E.~Yvinec, M.~Casbon, E.~Pot, I.~Penchev, G.~Liu,
  F.~Visin, K.~Kenealy, L.~Beyer, X.~Zhai, A.~Tsitsulin, R.~Busa-Fekete,
  A.~Feng, N.~Sachdeva, B.~Coleman, Y.~Gao, B.~Mustafa, I.~Barr, E.~Parisotto,
  D.~Tian, M.~Eyal, C.~Cherry, J.-T. Peter, D.~Sinopalnikov, S.~Bhupatiraju,
  R.~Agarwal, M.~Kazemi, D.~Malkin, R.~Kumar, D.~Vilar, I.~Brusilovsky, J.~Luo,
  A.~Steiner, A.~Friesen, A.~Sharma, A.~Sharma, A.~M. Gilady, A.~Goedeckemeyer,
  A.~Saade, A.~Feng, A.~Kolesnikov, A.~Bendebury, A.~Abdagic, A.~Vadi,
  A.~György, A.~S. Pinto, A.~Das, A.~Bapna, A.~Miech, A.~Yang, A.~Paterson,
  A.~Shenoy, A.~Chakrabarti, B.~Piot, B.~Wu, B.~Shahriari, B.~Petrini, C.~Chen,
  C.~L. Lan, C.~A. Choquette-Choo, C.~Carey, C.~Brick, D.~Deutsch, D.~Eisenbud,
  D.~Cattle, D.~Cheng, D.~Paparas, D.~S. Sreepathihalli, D.~Reid, D.~Tran,
  D.~Zelle, E.~Noland, E.~Huizenga, E.~Kharitonov, F.~Liu, G.~Amirkhanyan,
  G.~Cameron, H.~Hashemi, H.~Klimczak-Plucińska, H.~Singh, H.~Mehta, H.~T.
  Lehri, H.~Hazimeh, I.~Ballantyne, I.~Szpektor, I.~Nardini, J.~Pouget-Abadie,
  J.~Chan, J.~Stanton, J.~Wieting, J.~Lai, J.~Orbay, J.~Fernandez, J.~Newlan,
  J.~yeong Ji, J.~Singh, K.~Black, K.~Yu, K.~Hui, K.~Vodrahalli, K.~Greff,
  L.~Qiu, M.~Valentine, M.~Coelho, M.~Ritter, M.~Hoffman, M.~Watson,
  M.~Chaturvedi, M.~Moynihan, M.~Ma, N.~Babar, N.~Noy, N.~Byrd, N.~Roy,
  N.~Momchev, N.~Chauhan, N.~Sachdeva, O.~Bunyan, P.~Botarda, P.~Caron, P.~K.
  Rubenstein, P.~Culliton, P.~Schmid, P.~G. Sessa, P.~Xu, P.~Stanczyk,
  P.~Tafti, R.~Shivanna, R.~Wu, R.~Pan, R.~Rokni, R.~Willoughby, R.~Vallu,
  R.~Mullins, S.~Jerome, S.~Smoot, S.~Girgin, S.~Iqbal, S.~Reddy, S.~Sheth,
  S.~Põder, S.~Bhatnagar, S.~R. Panyam, S.~Eiger, S.~Zhang, T.~Liu,
  T.~Yacovone, T.~Liechty, U.~Kalra, U.~Evci, V.~Misra, V.~Roseberry,
  V.~Feinberg, V.~Kolesnikov, W.~Han, W.~Kwon, X.~Chen, Y.~Chow, Y.~Zhu,
  Z.~Wei, Z.~Egyed, V.~Cotruta, M.~Giang, P.~Kirk, A.~Rao, K.~Black, N.~Babar,
  J.~Lo, E.~Moreira, L.~G. Martins, O.~Sanseviero, L.~Gonzalez, Z.~Gleicher,
  T.~Warkentin, V.~Mirrokni, E.~Senter, E.~Collins, J.~Barral, Z.~Ghahramani,
  R.~Hadsell, Y.~Matias, D.~Sculley, S.~Petrov, N.~Fiedel, N.~Shazeer,
  O.~Vinyals, J.~Dean, D.~Hassabis, K.~Kavukcuoglu, C.~Farabet, E.~Buchatskaya,
  J.-B. Alayrac, R.~Anil, Dmitry, Lepikhin, S.~Borgeaud, O.~Bachem, A.~Joulin,
  A.~Andreev, C.~Hardin, R.~Dadashi, and L.~Hussenot.
\newblock Gemma 3 technical report, 2025.

\bibitem[Uschmajew and Vandereycken(2022)]{uschmajew2022note}
A.~Uschmajew and B.~Vandereycken.
\newblock A note on the optimal convergence rate of descent methods with fixed
  step sizes for smooth strongly convex functions.
\newblock \emph{Journal of Optimization Theory and Applications}, 194\penalty0
  (1):\penalty0 364--373, 2022.
\newblock \doi{10.1007/s10957-022-02032-z}.

\bibitem[Wang et~al.(2023)Wang, Wei, Schuurmans, Le, Chi, Narang, Chowdhery,
  and Zhou]{selfconsistency}
X.~Wang, J.~Wei, D.~Schuurmans, Q.~V. Le, E.~H. Chi, S.~Narang, A.~Chowdhery,
  and D.~Zhou.
\newblock Self-consistency improves chain of thought reasoning in language
  models.
\newblock In \emph{The Eleventh International Conference on Learning
  Representations}, 2023.

\bibitem[Wu et~al.(2026)Wu, He, Hu, Wei, Li, Yang, and Zhu]{criticL}
Y.~Wu, Y.~He, Z.~Hu, L.~Wei, R.~Li, Q.~Yang, and T.~Zhu.
\newblock {CritICL}: Inference-time weak-to-strong generalization from small
  language model failure modes, 2026.

\bibitem[Yang et~al.(2025)Yang, Li, Yang, Zhang, Hui, Zheng, Yu, Gao, Huang,
  Lv, Zheng, Liu, Zhou, Huang, Hu, Ge, Wei, Lin, Tang, Yang, Tu, Zhang, Yang,
  Yang, Zhou, Zhou, Lin, Dang, Bao, Yang, Yu, Deng, Li, Xue, Li, Zhang, Wang,
  Zhu, Men, Gao, Liu, Luo, Li, Tang, Yin, Ren, Wang, Zhang, Ren, Fan, Su,
  Zhang, Zhang, Wan, Liu, Wang, Cui, Zhang, Zhou, and Qiu]{qwen3}
A.~Yang, A.~Li, B.~Yang, B.~Zhang, B.~Hui, B.~Zheng, B.~Yu, C.~Gao, C.~Huang,
  C.~Lv, C.~Zheng, D.~Liu, F.~Zhou, F.~Huang, F.~Hu, H.~Ge, H.~Wei, H.~Lin,
  J.~Tang, J.~Yang, J.~Tu, J.~Zhang, J.~Yang, J.~Yang, J.~Zhou, J.~Zhou,
  J.~Lin, K.~Dang, K.~Bao, K.~Yang, L.~Yu, L.~Deng, M.~Li, M.~Xue, M.~Li,
  P.~Zhang, P.~Wang, Q.~Zhu, R.~Men, R.~Gao, S.~Liu, S.~Luo, T.~Li, T.~Tang,
  W.~Yin, X.~Ren, X.~Wang, X.~Zhang, X.~Ren, Y.~Fan, Y.~Su, Y.~Zhang, Y.~Zhang,
  Y.~Wan, Y.~Liu, Z.~Wang, Z.~Cui, Z.~Zhang, Z.~Zhou, and Z.~Qiu.
\newblock Qwen3 technical report, 2025.

\bibitem[Yu et~al.(2026{\natexlab{a}})Yu, Lin, Liu, Xu, Xu, Zhou, and
  Gao]{W2SOPD}
F.~Yu, Z.~Lin, X.~Liu, W.~Xu, M.~Xu, T.~Zhou, and J.~Gao.
\newblock Weak-to-strong on-policy distillation, 2026{\natexlab{a}}.

\bibitem[Yu et~al.(2025)Yu, Zhang, Zhu, Yuan, Zuo, Yue, Dai, Fan, Liu, Liu,
  Liu, Lin, Lin, Ma, Sheng, Tong, Zhang, Zhang, Zhang, Zhu, Zhu, Chen, Chen,
  Wang, Yu, Song, Wei, Zhou, Liu, Ma, Zhang, Yan, Qiao, Wu, and Wang]{dapo}
Q.~Yu, Z.~Zhang, R.~Zhu, Y.~Yuan, X.~Zuo, Y.~Yue, W.~Dai, T.~Fan, G.~Liu,
  L.~Liu, X.~Liu, H.~Lin, Z.~Lin, B.~Ma, G.~Sheng, Y.~Tong, C.~Zhang, M.~Zhang,
  W.~Zhang, H.~Zhu, J.~Zhu, J.~Chen, J.~Chen, C.~Wang, H.~Yu, Y.~Song, X.~Wei,
  H.~Zhou, J.~Liu, W.-Y. Ma, Y.-Q. Zhang, L.~Yan, M.~Qiao, Y.~Wu, and M.~Wang.
\newblock Dapo: An open-source llm reinforcement learning system at scale,
  2025.

\bibitem[Yu et~al.(2026{\natexlab{b}})Yu, Yang, Wang, Zhao, Jenkins, Zhang,
  Bansal, Yao, and Zhang]{icpo}
T.~Yu, Y.~Yang, Z.~Wang, K.~Zhao, P.~Jenkins, X.~Zhang, C.~Bansal, H.~Yao, and
  W.~Zhang.
\newblock Provable and practical in-context policy optimization for
  self-improvement.
\newblock In \emph{The Fourteenth International Conference on Learning
  Representations}, 2026{\natexlab{b}}.

\bibitem[Yuan et~al.(2023)Yuan, Yuan, Li, Dong, Lu, Tan, Zhou, and Zhou]{rft}
Z.~Yuan, H.~Yuan, C.~Li, G.~Dong, K.~Lu, C.~Tan, C.~Zhou, and J.~Zhou.
\newblock Scaling relationship on learning mathematical reasoning with large
  language models, 2023.

\bibitem[Zelikman et~al.(2022)Zelikman, Wu, Mu, and Goodman]{star}
E.~Zelikman, Y.~Wu, J.~Mu, and N.~Goodman.
\newblock {STaR}: Bootstrapping reasoning with reasoning.
\newblock In \emph{Advances in Neural Information Processing Systems},
  volume~35, 2022.
\newblock \doi{10.52202/068431-1126}.

\bibitem[Zhang and Math-AI(2024)]{aime24}
Y.~Zhang and T.~Math-AI.
\newblock American invitational mathematics examination (aime) 2024, 2024.

\bibitem[Zhang and Math-AI(2025)]{aime25}
Y.~Zhang and T.~Math-AI.
\newblock American invitational mathematics examination (aime) 2025, 2025.

\bibitem[Zhang et~al.(2026)Zhang, Khalifa, Zhang, Liu, Lee, Zhang, Bayat, and
  Wang]{zhang2025thinklogit}
Y.~Zhang, M.~Khalifa, L.~Zhang, X.~Liu, A.~Lee, X.~F. Zhang, F.~F. Bayat, and
  L.~Wang.
\newblock Logit arithmetic elicits long reasoning capabilities without
  training, 2026.

\bibitem[Zhou et~al.(2023)Zhou, Lu, Mishra, Brahma, Basu, Luan, Zhou, and
  Hou]{IFeval}
J.~Zhou, T.~Lu, S.~Mishra, S.~Brahma, S.~Basu, Y.~Luan, D.~Zhou, and L.~Hou.
\newblock Instruction-following evaluation for large language models, 2023.

\bibitem[Zhou et~al.(2024)Zhou, Liu, Liu, Dong, Yang, and Qiao]{W2SSearch}
Z.~Zhou, Z.~Liu, J.~Liu, Z.~Dong, C.~Yang, and Y.~Qiao.
\newblock Weak-to-strong search: Align large language models via searching over
  small language models.
\newblock In A.~Globerson, L.~Mackey, D.~Belgrave, A.~Fan, U.~Paquet,
  J.~Tomczak, and C.~Zhang, editors, \emph{Advances in Neural Information
  Processing Systems}, volume~37, pages 4819--4851. Curran Associates, Inc.,
  2024.
\newblock \doi{10.52202/079017-0156}.

\bibitem[Zuo et~al.(2025)Zuo, Zhang, Sheng, Qu, Cui, Zhu, Li, Zhang, Long, Hua,
  Qi, Sun, Ma, Yuan, Ding, and Zhou]{ttrl}
Y.~Zuo, K.~Zhang, L.~Sheng, S.~Qu, G.~Cui, X.~Zhu, H.~Li, Y.~Zhang, X.~Long,
  E.~Hua, B.~Qi, Y.~Sun, Z.~Ma, L.~Yuan, N.~Ding, and B.~Zhou.
\newblock {TTRL}: Test-time reinforcement learning.
\newblock In \emph{Advances in Neural Information Processing Systems},
  volume~38, 2025.
\newblock \doi{10.52202/085713-4376}.

\end{thebibliography}
\bibliographystyle{abbrvnat}

\clearpage
\appendix

\section{Proofs for Theoretical Justification}
\label{app:tracking}

We give an explicit version of \Cref{thm:opd_tracking}, state the supporting lemmas, and assemble the main proof before proving the lemmas. We then establish local curvature and quantify approximation error when coverage is partial.

\subsection{Quantitative Tracking Statement}
\label{app:tracking_statement}

We use the linear-softmax model and fixed prefix weights from \Cref{sec:theory}. All logits are finite. Let $\mathbf{M}=\operatorname{diag}(\mu_1,\ldots,\mu_n)$ and $\mathbf{\Sigma}=\mathbf{A}^\top \mathbf{M} \mathbf{A}$, with
\begin{align}
    \Lambda&=\lambda_{\max}(\mathbf{\Sigma}),\qquad
    \gamma=\lambda_{\min}^{+}(\mathbf{\Sigma})>0,
    \notag %
\end{align}
where $\lambda_{\min}^{+}$ denotes the smallest strictly positive eigenvalue. The identifiable parameter subspace is
\begin{align}
    \mathcal{U}&=\{\mathbf{E}\in\mathbb{R}^{d_S\times V}:\mathbf{A}^\dagger \mathbf{A} \mathbf{E}=\mathbf{E},\ \mathbf{E}\boldsymbol{1}=0\}.
    \notag %
\end{align}
Its columns lie in $\operatorname{range}(\mathbf{A}^\top)$, and its rows are centered across tokens. Define the averaged conditional objective
\begin{align}
    \mathcal{J}(\mathbf{W})&=\sum_i\mu_i\left[\langle \boldsymbol{p}_i(\mathbf{W}),\mathbf{\Delta}_{i,:}^\top\rangle
    -\beta D_{\mathrm{KL}}(\boldsymbol{p}_i(\mathbf{W})\,\|\,\boldsymbol{p}_i(\mathbf{W}_0))\right].
    \notag %
\end{align}
For arbitrary conditional distributions $\mathbf{P}=(\boldsymbol{p}_i)_i$, write $\mathcal{J}(\mathbf{P})$ for the same expression with $\boldsymbol{p}_i(\mathbf{W})$ replaced by $\boldsymbol{p}_i$.

\begin{proposition}[Explicit tracking rate]
\label{prop:tracking_rate}
Under the coverage condition of \Cref{thm:opd_tracking}, let $\mathbf{W}^\star=\mathbf{W}_0+\mathbf{A}^\dagger \mathbf{D}/\beta$ and define
\begin{align}
R_0 &= \|\mathbf{W}_0-\mathbf{W}^\star\|_F,\qquad a_{\max}=\max_i\|\boldsymbol{a}_i\|_2, \qquad m_\star =\min_{i,a}q_i(a)>0,\qquad
m=m_\star e^{-2a_{\max}R_0}.
\notag %
\end{align}
Then $\mathcal{J}(\mathbf{W}^\star)-\mathcal{J}(\mathbf{W})=\mathcal{L}(\mathbf{W})$ for every $\mathbf{W}$, and every finite stationary point induces the policy $(\boldsymbol{q}_i)_i$. The head $\mathbf{W}^\star$ is the unique minimizer in $\mathbf{W}_0+\mathcal{U}$.
For $0<\eta\le(\beta\Lambda)^{-1}$, gradient descent on $\mathcal{L}$ from $\mathbf{W}_0$ stays in this affine subspace and satisfies
\begin{align}
    \|\mathbf{W}_k-\mathbf{W}^\star\|_F&\le\rho^kR_0,
    &\rho&=1-\eta\beta m\gamma\in[0,1),\label{eq:tracking_rate_explicit}\\
    \sum_i\mu_iD_{\mathrm{KL}}(\boldsymbol{p}_i(\mathbf{W}_k)\,\|\,\boldsymbol{q}_i)
    &\le\frac{\Lambda R_0^2}{4}\rho^{2k}. \notag
\end{align}
\end{proposition}

\subsection{Supporting Lemmas}
\label{app:tracking_lemmas}

The first lemma identifies the inference-time target exactly. The second expresses its optimization error through the softmax Fisher matrix. The third controls this operator on the directions that change the student policy.

\begin{lemma}[Softmax centering and the objective gap]
\label[lemma]{lem:tracking_softmax}
For any logit row $\boldsymbol{z}$,
\begin{align}
    \log\operatorname{softmax}(\boldsymbol{z})
    &=\boldsymbol{z}-\operatorname{LSE}(\boldsymbol{z})\boldsymbol{1}^\top,
    \qquad \operatorname{LSE}(\boldsymbol{z})=\log\sum_a e^{z_a}.
    \notag %
\end{align}
Consequently, $\mathbf{D}=\mathbf{\Delta} \mathbf{C}=\mathbf{B}(\mathbf{\Theta}_+-\mathbf{\Theta}_-)\mathbf{C}$, and the state-wise maximizer is $\boldsymbol{q}_i=\operatorname{softmax}(\mathbf{W}_0^\top \boldsymbol{a}_i+\mathbf{D}_{i,:}^\top/\beta)$. For every collection of conditional distributions $\mathbf{P}=(\boldsymbol{p}_i)_i$,
\begin{align}
    \mathcal{J}(\mathbf{Q})-\mathcal{J}(\mathbf{P})
    &=\beta\sum_i\mu_iD_{\mathrm{KL}}(\boldsymbol{p}_i\,\|\,\boldsymbol{q}_i),
    \qquad \mathbf{Q}=(\boldsymbol{q}_i)_i.
    \label{eq:tracking_gap}
\end{align}
\end{lemma}

\begin{lemma}[Exact gradient and identifiable directions]
\label[lemma]{lem:tracking_gradient}
Suppose $\mathbf{W}^\star$ realizes $\mathbf{Q}$. Define
\begin{align}
    \mathbf{F}_i(\mathbf{W})&=\operatorname{diag}(\boldsymbol{p}_i(\mathbf{W}))-\boldsymbol{p}_i(\mathbf{W})\boldsymbol{p}_i(\mathbf{W})^\top,
    \qquad
    \mathcal{G}_{\mathbf{W}}[\mathbf{E}]=\sum_i\mu_i \boldsymbol{a}_i \boldsymbol{a}_i^\top \mathbf{E} \mathbf{F}_i(\mathbf{W}).
    \notag %
\end{align}
The operator $\mathcal{G}_{\mathbf{W}}$ is self-adjoint and positive semidefinite, and
\begin{align}
    \nabla\mathcal{L}(\mathbf{W})&=\beta\mathcal{G}_{\mathbf{W}}[\mathbf{W}-\mathbf{W}^\star],\label{eq:tracking_exact_gradient}\\
    \langle \mathbf{E},\mathcal{G}_{\mathbf{W}}[\mathbf{E}]\rangle_F
    &=\sum_i\mu_i\operatorname{Var}_{a\sim \boldsymbol{p}_i(\mathbf{W})}[(\boldsymbol{a}_i^\top \mathbf{E})_a].
    \label{eq:tracking_variance}
\end{align}
Its nullspace is $\{\mathbf{E}:\mathbf{A}\mathbf{E}\mathbf{C}=0\}$, and its range lies in $\mathcal{U}$.
\end{lemma}

\begin{lemma}[Restricted Fisher bounds and policy error]
\label[lemma]{lem:tracking_curvature}
For every $\mathbf{W}$, $0\preceq\mathcal{G}_{\mathbf{W}}\preceq(\Lambda/2)\mathbf{I}$. If $\min_{i,a}[\boldsymbol{p}_i(\mathbf{W})]_a\ge\varepsilon>0$, then for all $\mathbf{E}\in\mathcal{U}$,
\begin{align}
    \varepsilon\gamma\|\mathbf{E}\|_F^2
    &\le\langle \mathbf{E},\mathcal{G}_{\mathbf{W}}[\mathbf{E}]\rangle_F
    \le\frac{\Lambda}{2}\|\mathbf{E}\|_F^2.
    \notag %
\end{align}
For any two logit vectors $\boldsymbol{z},\widetilde{\boldsymbol{z}}\in\mathbb{R}^V$,
\begin{align}
    D_{\mathrm{KL}}(\operatorname{softmax}(\boldsymbol{z})\,\|\,\operatorname{softmax}(\widetilde{\boldsymbol{z}}))
    &\le\frac14\|\boldsymbol{z}-\widetilde{\boldsymbol{z}}\|_2^2.
    \label{eq:tracking_logit_kl}
\end{align}
\end{lemma}

\subsection{Proof of the Main Theorem}
\label{app:tracking_main_proof}

\begin{proof}[Proof of \Cref{thm:opd_tracking,prop:tracking_rate}]
\textbf{Realizability.}
Two finite logit vectors have the same softmax if and only if they differ by a multiple of $\boldsymbol{1}$. By \Cref{lem:tracking_softmax}, a head $\mathbf{W}$ realizes $\mathbf{Q}$ exactly when
\begin{align}
    \mathbf{A}(\mathbf{W}-\mathbf{W}_0)\mathbf{C}&=\mathbf{D}/\beta.
    \label{eq:tracking_realization_equation}
\end{align}
Such a head exists if and only if every column of $\mathbf{D}$ belongs to $\operatorname{col}(\mathbf{A})$, equivalently $(\mathbf{I}-\mathbf{A}\mathbf{A}^\dagger)\mathbf{D}=0$. Under this condition, the choice $\mathbf{W}^\star=\mathbf{W}_0+\mathbf{A}^\dagger \mathbf{D}/\beta$ satisfies \Cref{eq:tracking_realization_equation}. Substituting $\mathbf{P}=\mathbf{P}_{\mathbf{W}}$ into \Cref{eq:tracking_gap} gives $\mathcal{J}(\mathbf{W}^\star)-\mathcal{J}(\mathbf{W})=\mathcal{L}(\mathbf{W})$.

\textbf{Identifiable directions and stationary points.}
The initial error $\mathbf{E}_0=\mathbf{W}_0-\mathbf{W}^\star=-\mathbf{A}^\dagger \mathbf{D}/\beta$ lies in $\mathcal{U}$ because $\mathbf{D}\mathbf{C}=\mathbf{D}$. By \Cref{lem:tracking_gradient}, every subsequent gradient also lies in $\mathcal{U}$. Therefore $\mathbf{E}_k=\mathbf{W}_k-\mathbf{W}^\star\in\mathcal{U}$ for all $k$.
At any finite stationary head $\mathbf{W}$, \Cref{eq:tracking_exact_gradient,eq:tracking_variance} imply
\begin{align}
    0&=\langle \mathbf{W}-\mathbf{W}^\star,\nabla\mathcal{L}(\mathbf{W})\rangle_F
    =\beta\sum_i\mu_i\operatorname{Var}_{a\sim \boldsymbol{p}_i(\mathbf{W})}[(\boldsymbol{a}_i^\top(\mathbf{W}-\mathbf{W}^\star))_a]. \notag
\end{align}
Every $\mu_i$ and every softmax probability are positive. Each variance is therefore zero, so $\mathbf{A}(\mathbf{W}-\mathbf{W}^\star)\mathbf{C}=0$ and $\boldsymbol{p}_i(\mathbf{W})=\boldsymbol{q}_i$ for every $i$. On $\mathcal{U}$ this nullspace contains only zero: if $\mathbf{E}\mathbf{C}=\mathbf{E}$, $\mathbf{A}\mathbf{E}=0$, and $\mathbf{A}^\dagger \mathbf{A}\mathbf{E}=\mathbf{E}$, then $\mathbf{E}=0$. This proves uniqueness in $\mathbf{W}_0+\mathcal{U}$.

\textbf{A bounded gradient trajectory.}
The exact gradient identity gives
\begin{align}
    \mathbf{E}_{k+1}&=(\mathbf{I}-\eta\beta\mathcal{G}_{\mathbf{W}_k})\mathbf{E}_k.
    \label{eq:tracking_error_recursion}
\end{align}
The upper bound in \Cref{lem:tracking_curvature} and $\eta\beta\Lambda\le1$ imply that this operator has all eigenvalues in $[0,1]$. Hence $\|\mathbf{E}_k\|_F\le R_0$ for every $k$. For each prefix,
\begin{align}
    \|\boldsymbol{a}_i^\top \mathbf{E}_k\|_\infty&\le\|\boldsymbol{a}_i\|_2R_0\le a_{\max}R_0. \notag
\end{align}
Writing $\boldsymbol{e}_i=\boldsymbol{a}_i^\top \mathbf{E}_k$, the current policy is an exponential tilt of $\boldsymbol{q}_i$:
\begin{align}
    [\boldsymbol{p}_i(\mathbf{W}_k)]_a
    &=\frac{q_i(a)e^{e_i(a)}}{\sum_bq_i(b)e^{e_i(b)}}
    \ge m_\star e^{-2a_{\max}R_0}=m. \notag
\end{align}

\textbf{Geometric tracking.}
The lower bound in \Cref{lem:tracking_curvature} now holds uniformly along the iterates. Since $\mathbf{E}_k\in\mathcal{U}$, \Cref{eq:tracking_error_recursion} gives
\begin{align}
    \|\mathbf{E}_{k+1}\|_F&\le(1-\eta\beta m\gamma)\|\mathbf{E}_k\|_F. \notag
\end{align}
Iteration proves \Cref{eq:tracking_rate_explicit}. Finally, \Cref{eq:tracking_logit_kl} yields
\begin{align}
    \sum_i\mu_iD_{\mathrm{KL}}(\boldsymbol{p}_i(\mathbf{W}_k)\,\|\,\boldsymbol{q}_i)
    &\le\frac14\sum_i\mu_i\|\boldsymbol{a}_i^\top \mathbf{E}_k\|_2^2 \notag\\
    &=\frac14\operatorname{tr}(\mathbf{E}_k^\top\mathbf{\Sigma} \mathbf{E}_k)
    \le\frac{\Lambda}{4}\|\mathbf{E}_k\|_F^2
    \le\frac{\Lambda R_0^2}{4}\rho^{2k}. \notag
\end{align}
This proves both the parameter and policy bounds. When $\mathbf{D}=0$, $\mathbf{W}^\star=\mathbf{W}_0$ and the iterates already equal the target.
\end{proof}

\subsection{Proofs of the Supporting Lemmas}
\label{app:tracking_lemma_proofs}

\begin{proof}[Proof of \Cref{lem:tracking_softmax}]
The log-softmax identity follows from its normalization. Multiplication by $\mathbf{C}$ removes the rowwise constant, giving $\mathbf{\Delta} \mathbf{C}=\mathbf{B}(\mathbf{\Theta}_+-\mathbf{\Theta}_-)\mathbf{C}$.
For each prefix, define $Z_i=\sum_a [\boldsymbol{p}_i(\mathbf{W}_0)]_a\exp(\Delta_{ia}/\beta)$. The Gibbs policy satisfies
\begin{align}
    \log q_i(a)&=\log [\boldsymbol{p}_i(\mathbf{W}_0)]_a+\Delta_{ia}/\beta-\log Z_i. \notag
\end{align}
Substituting into the KL divergence gives, for every $\boldsymbol{p}_i$,
\begin{align}
    \langle \boldsymbol{p}_i,\mathbf{\Delta}_{i,:}^\top\rangle
    -\beta D_{\mathrm{KL}}(\boldsymbol{p}_i\,\|\,\boldsymbol{p}_i(\mathbf{W}_0))
    &=\beta\log Z_i-\beta D_{\mathrm{KL}}(\boldsymbol{p}_i\,\|\,\boldsymbol{q}_i). \notag
\end{align}
At $\boldsymbol{p}_i=\boldsymbol{q}_i$ the last term is zero. Averaging over prefixes proves \Cref{eq:tracking_gap} and establishes the claimed optimizer.
\end{proof}

\begin{proof}[Proof of \Cref{lem:tracking_gradient}]
For $\boldsymbol{p}=\operatorname{softmax}(\boldsymbol{z})$, its Jacobian is $\mathbf{F}(\boldsymbol{p})=\operatorname{diag}(\boldsymbol{p})-\boldsymbol{p}\boldsymbol{p}^\top$, with $\mathbf{F}(\boldsymbol{p})\boldsymbol{1}=0$. This is also the Fisher information with respect to logits:
\begin{align}
    \mathbf{F}(\boldsymbol{p})&=\mathbb{E}_{a\sim \boldsymbol{p}}\left[(\boldsymbol{e}_a-\boldsymbol{p})(\boldsymbol{e}_a-\boldsymbol{p})^\top\right]
    =\mathbb{E}_{a\sim \boldsymbol{p}}\left[\nabla_{\boldsymbol{z}}\log p(a)\nabla_{\boldsymbol{z}}\log p(a)^\top\right], \notag
\end{align}
where $\boldsymbol{e}_a$ is the coordinate vector for token $a$. Differentiating reverse KL gives
\begin{align}
    \nabla_{\boldsymbol{z}}D_{\mathrm{KL}}(\boldsymbol{p}\,\|\,\boldsymbol{q})&=\mathbf{F}(\boldsymbol{p})(\log \boldsymbol{p}-\log \boldsymbol{q}). \notag
\end{align}
For $\boldsymbol{q}_i=\boldsymbol{p}_i(\mathbf{W}^\star)$, the log-probability difference equals $(\mathbf{W}-\mathbf{W}^\star)^\top \boldsymbol{a}_i$ up to a multiple of $\boldsymbol{1}$. The chain rule through $\boldsymbol{z}_i=\mathbf{W}^\top \boldsymbol{a}_i$ therefore gives
\begin{align}
    \nabla_{\mathbf{W}}D_{\mathrm{KL}}(\boldsymbol{p}_i(\mathbf{W})\,\|\,\boldsymbol{q}_i)
    &=\boldsymbol{a}_i \boldsymbol{a}_i^\top(\mathbf{W}-\mathbf{W}^\star)\mathbf{F}_i(\mathbf{W}). \notag
\end{align}
Summing proves \Cref{eq:tracking_exact_gradient}.

Both $\boldsymbol{a}_i \boldsymbol{a}_i^\top$ and $\mathbf{F}_i(\mathbf{W})$ are symmetric, so $\mathcal{G}_{\mathbf{W}}$ is self-adjoint under the Frobenius inner product. Its quadratic form is
\begin{align}
    \langle \mathbf{E},\mathcal{G}_{\mathbf{W}}[\mathbf{E}]\rangle_F
    &=\sum_i\mu_i(\boldsymbol{a}_i^\top \mathbf{E})\mathbf{F}_i(\mathbf{W})(\boldsymbol{a}_i^\top \mathbf{E})^\top, \notag
\end{align}
which is the nonnegative variance in \Cref{eq:tracking_variance}. Since all probabilities and weights are positive, it vanishes exactly when every row of $\mathbf{A}\mathbf{E}$ is constant across actions, or $\mathbf{A}\mathbf{E}\mathbf{C}=0$. Each operator output has columns in $\operatorname{range}(\mathbf{A}^\top)$ and satisfies $\mathcal{G}_{\mathbf{W}}[\mathbf{E}]\boldsymbol{1}=0$, proving the range statement.
\end{proof}

\begin{proof}[Proof of \Cref{lem:tracking_curvature}]
For any $\boldsymbol{p}$ with $\min_a p_a\ge\varepsilon$ and any $\boldsymbol{v}\in\mathbb{R}^V$,
\begin{align}
    \boldsymbol{v}^\top \mathbf{F}(\boldsymbol{p})\boldsymbol{v}
    &=\min_c\sum_a p_a(v_a-c)^2
    \ge\varepsilon\min_c\sum_a(v_a-c)^2
    =\varepsilon\|\mathbf{C}\boldsymbol{v}\|_2^2. \notag
\end{align}
The absolute row sum of $\mathbf{F}(\boldsymbol{p})$ is $2p_a(1-p_a)\le1/2$, so $\mathbf{F}(\boldsymbol{p})\preceq \mathbf{I}/2$. Applying these bounds to $\boldsymbol{v}=(\boldsymbol{a}_i^\top \mathbf{E})^\top$ gives
\begin{align}
    \langle \mathbf{E},\mathcal{G}_{\mathbf{W}}[\mathbf{E}]\rangle_F
    &\le\frac12\operatorname{tr}(\mathbf{E}^\top\mathbf{\Sigma} \mathbf{E})
    \le\frac{\Lambda}{2}\|\mathbf{E}\|_F^2. \notag
\end{align}
For $\mathbf{E}\in\mathcal{U}$, the rows of $\mathbf{A}\mathbf{E}$ are centered, and its parameter columns lie in $\operatorname{range}(\mathbf{\Sigma})=\operatorname{range}(\mathbf{A}^\top)$. Therefore
\begin{align}
    \langle \mathbf{E},\mathcal{G}_{\mathbf{W}}[\mathbf{E}]\rangle_F
    &\ge\varepsilon\operatorname{tr}(\mathbf{E}^\top\mathbf{\Sigma} \mathbf{E})
    \ge\varepsilon\gamma\|\mathbf{E}\|_F^2. \notag
\end{align}
Only positive eigenvalues of $\mathbf{\Sigma}$ enter this bound.

For the final claim, the Hessian of $\operatorname{LSE}$ is $\mathbf{F}(\operatorname{softmax}(\boldsymbol{z}))\preceq \mathbf{I}/2$. Its Bregman-divergence identity gives
\begin{align}
    D_{\mathrm{KL}}(\operatorname{softmax}(\boldsymbol{z})\,\|\,\operatorname{softmax}(\widetilde{\boldsymbol{z}}))
    &=\operatorname{LSE}(\widetilde{\boldsymbol{z}})-\operatorname{LSE}(\boldsymbol{z})
      -\langle\operatorname{softmax}(\boldsymbol{z}),\widetilde{\boldsymbol{z}}-\boldsymbol{z}\rangle \notag\\
    &\le\tfrac14\|\widetilde{\boldsymbol{z}}-\boldsymbol{z}\|_2^2. \notag
\end{align}
This completes the proof.
\end{proof}

\subsection{Local Curvature and Other Optimization Methods}
\label{app:tracking_local_curvature}

The exact gradient factorization also identifies the local geometry at the target head.

\begin{proposition}[Local strong convexity on identifiable directions]
\label{prop:tracking_local_curvature}
Under the coverage condition, the Hessian at $\mathbf{W}^\star$, restricted to $\mathcal{U}$, satisfies
\begin{align}
    \beta m_\star\gamma \mathbf{I}_{\mathcal{U}}
    &\preceq\nabla^2\mathcal{L}(\mathbf{W}^\star)|_{\mathcal{U}}
    \preceq\frac{\beta\Lambda}{2}\mathbf{I}_{\mathcal{U}}. \notag
\end{align}
Consequently, there is a neighborhood of $\mathbf{W}^\star$ in $\mathbf{W}_0+\mathcal{U}$ on which $\mathcal{L}$ is strongly convex and smooth.
\end{proposition}

\begin{proof}
Differentiate \Cref{eq:tracking_exact_gradient}. At $\mathbf{W}=\mathbf{W}^\star$, the term differentiating $\mathcal{G}_{\mathbf{W}}$ is multiplied by $\mathbf{W}-\mathbf{W}^\star=0$. Thus, for $\mathbf{H}\in\mathcal{U}$,
\begin{align}
    \nabla^2\mathcal{L}(\mathbf{W}^\star)[\mathbf{H}]
    &=\beta\mathcal{G}_{\mathbf{W}^\star}[\mathbf{H}]. \notag
\end{align}
Apply \Cref{lem:tracking_curvature} with $\varepsilon=m_\star$. The Hessian is continuous because all finite softmax probabilities are positive. A sufficiently small convex neighborhood therefore has restricted Hessian bounded below by $(\beta m_\star\gamma/2)\mathbf{I}_{\mathcal{U}}$ and above by $\beta\Lambda \mathbf{I}_{\mathcal{U}}$, proving the claim.
\end{proof}

In coordinates for $\mathcal{U}$, a fixed positive-definite preconditioner preserves this local optimization structure after a change of variables. Established variable-metric results control the resulting contraction through bounds on the preconditioner's eigenvalues \citep{uschmajew2022note}. Stochastic-gradient analyses additionally impose moment conditions on the gradient noise and suitable step-size schedules \citep{bottou2018optimization}. The local curvature above supplies the objective geometry for such extensions when their iterates remain in this neighborhood.

\subsection{Partial Coverage and Support-Gated Guidance}
\label{app:tracking_approximation}

For a centered guidance matrix $\mathbf{D}$, its component outside the student feature span controls the quality of the best realizable approximation. This statement applies directly to the exact reverse-KL loss in \Cref{eq:theory_loss}.

\begin{proposition}[Approximation under partial coverage]
\label{prop:tracking_approximation}
For any centered $\mathbf{D}$, let
\begin{align}
    \mathbf{A}_\mu&=\mathbf{M}^{1/2}\mathbf{A},\qquad \mathbf{D}_\mu=\mathbf{M}^{1/2}\mathbf{D},\qquad
    \mathbf{R}_\perp=(\mathbf{I}-\mathbf{A}_\mu \mathbf{A}_\mu^\dagger)\mathbf{D}_\mu.
    \label{eq:tracking_projection_residual}
\end{align}
Define $\boldsymbol{q}_i$ as in \Cref{sec:theory} and $\widehat{\mathbf{W}}=\mathbf{W}_0+\mathbf{A}_\mu^\dagger \mathbf{D}_\mu/\beta$. Then
\begin{align}
    \sum_i\mu_iD_{\mathrm{KL}}(\boldsymbol{p}_i(\widehat{\mathbf{W}})\,\|\,\boldsymbol{q}_i)
    &\le\frac{\|\mathbf{R}_\perp\|_F^2}{4\beta^2}.
    \label{eq:tracking_approximation}
\end{align}
More generally, if $\mathcal{L}(\widetilde{\mathbf{W}})-\inf_{\mathbf{W}}\mathcal{L}(\mathbf{W})\le\varepsilon_{\mathrm{opt}}$, then
\begin{align}
    \sum_i\mu_iD_{\mathrm{KL}}(\boldsymbol{p}_i(\widetilde{\mathbf{W}})\,\|\,\boldsymbol{q}_i)
    &\le\frac{\|\mathbf{R}_\perp\|_F^2}{4\beta^2}
       +\frac{\varepsilon_{\mathrm{opt}}}{\beta}.
    \label{eq:tracking_total_error}
\end{align}
\end{proposition}

\begin{proof}
The weighted logit displacement between $\widehat{\mathbf{W}}$ and the target is
\begin{align}
    \mathbf{M}^{1/2}\bigl(\mathbf{A}\widehat{\mathbf{W}}-\mathbf{A}\mathbf{W}_0-\mathbf{D}/\beta\bigr)
    &=-\mathbf{R}_\perp/\beta. \notag
\end{align}
Apply \Cref{eq:tracking_logit_kl} at every prefix and sum with weights $\mu_i$ to obtain \Cref{eq:tracking_approximation}. Since $\inf_{\mathbf{W}}\mathcal{L}(\mathbf{W})\le\mathcal{L}(\widehat{\mathbf{W}})$, the optimization-gap assumption gives \Cref{eq:tracking_total_error} after dividing by $\beta$.
\end{proof}

For support-aware steering, set $\mathbf{D}=\widetilde{\mathbf{\Delta}} \mathbf{C}$, with $\widetilde{\mathbf{\Delta}}$ defined in \Cref{eq:gated_steering_signal}. The residual in \Cref{eq:tracking_projection_residual} then measures how well the student features represent the gated guidance. The same formulation accommodates teacher pairs with different feature maps by using their actual centered logit difference $(\mathbf{Z}_+-\mathbf{Z}_-)\mathbf{C}$.

\subsection{Coverage, Rank, and the Transferred Signal}
\label{app:tracking_coverage}

\paragraph{Only the changed directions need coverage.}
With shared teacher features, $\mathbf{D}=\mathbf{B}(\mathbf{\Theta}_+-\mathbf{\Theta}_-)\mathbf{C}$. Thus
\begin{align}
    \operatorname{rank}(\mathbf{W}^\star-\mathbf{W}_0)&=\operatorname{rank}(\mathbf{D})
    \le\min\{\operatorname{rank}(\mathbf{B}),\operatorname{rank}(\mathbf{\Theta}_+-\mathbf{\Theta}_-),V-1\}. \notag
\end{align}
The equality follows from $\mathbf{A}\mathbf{A}^\dagger \mathbf{D}=\mathbf{D}$: multiplication by $\mathbf{A}^\dagger$ cannot increase rank, and multiplication back by $\mathbf{A}$ recovers $\mathbf{D}$. Coverage holds whenever $\operatorname{col}(\mathbf{B})\subseteq\operatorname{col}(\mathbf{A})$, and more generally whenever the particular changed directions in $\mathbf{D}$ lie in $\operatorname{col}(\mathbf{A})$. It is this alignment of prefix-dependent features that gives the weak-to-strong interpretation.

\paragraph{Positive support in the student representation.}
Let $\mathbf{A}_\mu=\mathbf{U}_r\mathbf{S}_r\mathbf{V}_r^\top$ be its compact singular value decomposition, with $\mathbf{S}_r=\operatorname{diag}(\sigma_1,\ldots,\sigma_r)$ and $\sigma_j>0$. Under exact coverage, write $\mathbf{D}_\mu=\mathbf{U}_r\mathbf{K}$. The canonical head change then satisfies
\begin{align}
    \mathbf{W}^\star-\mathbf{W}_0&=\beta^{-1}\mathbf{V}_r\mathbf{S}_r^{-1}\mathbf{K},
    \qquad
    \|\mathbf{W}^\star-\mathbf{W}_0\|_F^2
    =\beta^{-2}\sum_{j=1}^r\frac{\|\mathbf{K}_{j,:}\|_2^2}{\sigma_j^2}.
    \label{eq:tracking_signal_support}
\end{align}
Only nonzero singular directions are inverted. The rate in \Cref{prop:tracking_rate} uses $\gamma=\min_j\sigma_j^2$ on the identifiable subspace; \Cref{eq:tracking_signal_support} describes the size of the particular teacher-induced update. For a fixed guidance matrix and prefix distribution, expanding the student feature span decreases the projection residual in \Cref{eq:tracking_projection_residual}. Thus nested representational coverage improves the approximation bound in \Cref{prop:tracking_approximation}.

\paragraph{Contrastive transfer and general policy signals.}
The required signal depends on the target. Contrastive transfer needs $(\mathbf{I}-\mathbf{A}\mathbf{A}^\dagger)(\mathbf{Z}_+-\mathbf{Z}_-)\mathbf{C}=0$, whereas directly matching the positive teacher needs $(\mathbf{I}-\mathbf{A}\mathbf{A}^\dagger)\mathbf{Z}_+\mathbf{C}=0$. A shared teacher component can cancel in the first condition. For example, let $\boldsymbol{u}_1,\boldsymbol{u}_2,\boldsymbol{u}_3,\boldsymbol{u}_4$ be orthonormal prefix vectors, set $\mathbf{A}=[\boldsymbol{u}_1,\boldsymbol{u}_2,\boldsymbol{u}_3]$ and $\mathbf{B}=[\boldsymbol{u}_4,\boldsymbol{u}_1]$, and take nonzero centered token vectors $\boldsymbol{v},\boldsymbol{w}$. The teacher logits
\begin{align}
    \mathbf{Z}_-&=\boldsymbol{u}_4\boldsymbol{v}^\top,\qquad \mathbf{Z}_+=\boldsymbol{u}_4\boldsymbol{v}^\top+\boldsymbol{u}_1\boldsymbol{w}^\top \notag
\end{align}
have difference $\mathbf{D}=\boldsymbol{u}_1\boldsymbol{w}^\top\in\operatorname{col}(\mathbf{A})$. The student realizes this change exactly even though the full teacher retains the uncovered component $\boldsymbol{u}_4\boldsymbol{v}^\top$.

For any fully specified token signal $\mathbf{R}$, the corresponding state-wise Gibbs policy is $\operatorname{softmax}_{\mathrm{row}}(\mathbf{A}\mathbf{W}_0+\mathbf{R}\mathbf{C}/\beta)$, and its exact realization requires $(\mathbf{I}-\mathbf{A}\mathbf{A}^\dagger)\mathbf{R}\mathbf{C}=0$. If this condition fails, the optimal mean reverse-KL error is strictly positive. Indeed, a sequence of heads with KL error tending to zero would have probabilities converging to the strictly positive target at each prefix. Their centered log probabilities would then converge to the target's centered logits. Since all student centered logits belong to the closed linear space $\{\mathbf{A}\mathbf{U}\mathbf{C}:\mathbf{U}\in\mathbb{R}^{d_S\times V}\}$, this would imply the missing coverage condition. The teacher-change factorization identifies a structured signal for which coverage can be supplied by a richer student's features.

\section{Implementation Details}
\label{app:details}

\paragraph{Models and teacher training.}
Our main experiments initialize the student and reference policy from Qwen3-4B-Base and use Qwen3-1.7B-Base as the negative teacher $T^-$. For each domain, we independently train a positive teacher $T^+$ from Qwen3-1.7B-Base using GRPO. We use the step-400 checkpoints for math and code and the step-200 checkpoint for IF. \Cref{tab:teacher_training_config} summarizes the teacher training configurations.

In IF, the teacher's responses become shorter and less diverse between steps 200 and 400, even as its validation score increases. The median response length on IFEval decreases from 152 to 22 words, the fraction of responses shorter than five words increases from 8\% to 40\%, and token entropy decreases from 2.4 to 0.09. We use the step-200 checkpoint based on these response-quality measurements.

\begin{table}[ht]
    \centering
    \small
    \setlength{\tabcolsep}{6pt}
    \caption{
        \textbf{GRPO training configurations for positive teachers.}
        Each domain uses an independently trained Qwen3-1.7B-Base model.
    }
    \label{tab:teacher_training_config}
    \begin{tabular}{ll}
        \toprule
        \textbf{Hyperparameter} & \textbf{Value} \\
        \midrule
        Initialization & Qwen3-1.7B-Base \\
        Prompts per step & 32 \\
        Responses per prompt & 8 \\
        Learning rate & $1\times10^{-6}$ \\
        Policy clipping $\epsilon$ & 0.2 \\
        KL regularization & None \\
        Maximum response length & 7,168 \\
        Selected steps (Math/Code/IF) & 400 / 400 / 200 \\
        Number of GPUs & 4 \\
        \bottomrule
    \end{tabular}
\end{table}

\paragraph{On-policy distillation baselines.}
For Direct-OPD \citep{DirectOPD}, single-domain runs use 100 training steps, with 64 prompts and four responses per prompt at each step, totaling 6,400 prompt instances per domain. Mixed-domain runs use 300 steps, interleaving 6,400 prompt instances from each domain for a total of 19,200.

Single-domain Direct-OPD uses the corresponding domain-specific positive teacher and the shared negative teacher. Routed Direct-OPD selects the positive teacher by prompt domain during mixed-domain training. \Cref{tab:opd_hparams} lists the configurations for both variants.

We also evaluate $\Delta$-MOPD \citep{delta-mopd} in the mixed-domain setting and Waypoint-OPD \citep{waypointOPD} on math, following the training configurations in their respective papers. $\Delta$-MOPD aggregates contrastive signals from multiple domain-specific teachers. Waypoint-OPD uses an intermediate checkpoint to assess the consistency of policy shifts and gate the distillation signal. We use the step-200 and step-400 checkpoints of the same math GRPO run as its intermediate and final teachers.

\begin{table}[htbp]
    \centering
    \small
    \setlength{\tabcolsep}{6pt}
    \caption{
        \textbf{Training hyperparameters for Direct-OPD baselines.}
        These configurations are used for single-domain Direct-OPD
        and Routed Direct-OPD.
    }
    \label{tab:opd_hparams}
    \begin{tabular}{ll}
        \toprule
        \textbf{Hyperparameter} & \textbf{Value} \\
        \midrule
        Student initialization & Qwen3-4B-Base \\
        Prompts per step & 64 \\
        Responses per prompt & 4 \\
        Updates per step & 1 \\
        Optimizer & AdamW \\
        Learning rate & $1\times10^{-6}$ \\
        LR schedule & Constant \\
        Warmup & None \\
        Adam $\beta_1,\beta_2$ & 0.9, 0.999 \\
        Weight decay & 0.01 \\
        Gradient clipping & 1.0 \\
        Policy ratio clipping & 0.2 \\
        Loss reduction & Token mean \\
        Maximum prompt length & 1,024 \\
        Maximum response length & 7,168 \\
        Sampling temperature & 1.0 \\
        Top-$p$ & 1.0 \\
        Teacher scoring & Student top-16 \\
        Rollout engine & vLLM \\
        Actor precision & FP32 \\
        Hardware & 4 $\times$ RTX Pro 6000 \\
        \bottomrule
    \end{tabular}
\end{table}

\paragraph{Inference-time-guided rejection fine-tuning.}
For \methodname{}, we construct the gated steering policy from the domain-specific positive teacher, the shared negative teacher, and the reference student. We match the OPD prompt-instance budgets and generate one response per prompt instance: 6,400 responses per single-domain run and 19,200 in the mixed-domain setting. Mixed-domain RFT combines the retained trajectories from all three domains.

We retain completed trajectories that pass the task verifier and format constraints, then fine-tune the student on these trajectories with supervised learning. Unless otherwise specified, results use four epochs of RFT. \Cref{tab:steering_rft_hparams} lists the steering and fine-tuning configurations.

\begin{table}[htbp]
    \centering
    \small
    \setlength{\tabcolsep}{6pt}
    \caption{
        \textbf{Hyperparameters for \methodname{}.}
    }
    \label{tab:steering_rft_hparams}
    \begin{tabular}{ll}
        \toprule
        \textbf{Hyperparameter} & \textbf{Value} \\
        \midrule
        \multicolumn{2}{l}{\textit{Steering and rollout generation}} \\
        Support threshold $\tau$ & $10^{-4}$ \\
        Responses per prompt & 1 \\
        Sampling temperature & 1.0 \\
        Top-$p$ & 1.0 \\
        Maximum tokens (Math) & 8,192 \\
        Maximum tokens (Code/IF) & 4,096 \\
        Qwen3 decoding mode & Non-thinking \\
        \midrule
        \multicolumn{2}{l}{\textit{Rejection fine-tuning}} \\
        Student initialization & Qwen3-4B-Base \\
        Optimizer & AdamW \\
        Learning rate & $1\times10^{-5}$ \\
        LR schedule & Constant after warmup \\
        Warmup ratio & 0.05 \\
        Adam $\beta_1,\beta_2$ & 0.9, 0.95 \\
        Weight decay & 0 \\
        Gradient clipping & 1.0 \\
        Batch size & 64 sequences \\
        RFT epochs & 4 \\
        \bottomrule
    \end{tabular}
\end{table}

\paragraph{Evaluation protocol.}
Mathematical reasoning evaluations use avg@16 on AIME 2024, AIME 2025, and AMC 2023, and avg@4 on MATH500, Minerva, and OlympiadBench. All use a maximum generation length of 8,192 tokens.

For code, we report pooled avg@4 across HumanEval+ and MBPP+, with a maximum generation length of 4,096 tokens. For instruction following, we report strict prompt-level accuracy on IFEval, averaged over four generations per prompt with the same 4,096-token limit.

\paragraph{Validation sets and steering evaluation.}
For the steering analyses in \Cref{fig:gate_ablation,fig:ckpt_selection}, we use separate domain-specific validation sets rather than the test benchmarks used in the main experiments.

For math, we construct a validation set of 300 problems from the DAPO-Math-17k training pool (17,911 problems), excluding prompts used for \methodname{} rollout generation. For code, we use 670 problems from the held-out test split of code-r1-12k. For instruction following, we use 1,000 prompts from the Nemotron-RL-IF training pool (43,551 prompts), excluding prompts used for \methodname{} rollout generation.

For both analyses, we generate four responses per validation prompt with a sampling temperature of 0.7, top-$p$ of 0.95, and maximum generation lengths of 8,192 tokens for math and 4,096 tokens for code and instruction following. The steering coefficient is set to $\lambda=1$ ($\beta=1$). \Cref{fig:gate_ablation} compares steering with and without support-aware gating, while \Cref{fig:ckpt_selection} compares gated steering with Direct-OPD across teacher checkpoints.

These validation experiments are used to analyze steering behavior and teacher checkpoint selection. The main experimental comparisons use the separate test benchmarks and evaluation metrics described in \Cref{sec:exp_setup}.

\end{document}